\documentclass[manuscript,screen]{acmart}
\usepackage{glossaries}
\usepackage{svg}
\usepackage{preprint_header}

\newglossaryentry{tNLP}
{
    name=tNLP,
    description={traditional natural language processing}
}

\newglossaryentry{tbML}
{
    name=tbML,
    description={transformer-based machine learning}
}

\newglossaryentry{SVM}
{
    name=SVM,
    description={support vector machine}
}

\newglossaryentry{GCN}
{
    name=GCN,
    description={graph convolutional network}
}
\newglossaryentry{GRU}
{
    name=GRU,
    description={gated recurrent unit}
}

\newglossaryentry{HAN}
{
    name=HAN,
    description={hierarchical attention network}
}

\newglossaryentry{AttnBL}
{
    name=AttnBL,
    description={attention-based bidirectional long short-term memory}
}

\newglossaryentry{UMAP}
{
    name=UMAP,
    description={Uniform manifold approximation and projection for dimension reduction}
}

\newglossaryentry{RNN}
{
    name=RNN,
    description={recurrent neural network}
}

\newglossaryentry{RF}
{
    name=RF,
    description={random forest}
}

\newglossaryentry{ntbML}
{
    name=ntbML,
    description={non-transformer-based machine learning}
}

\newglossaryentry{NN}
{
    name=NN,
    description={neural network}
}

\newglossaryentry{nNN}
{
    name=nNN,
    description={non-neural networks}
}

\newglossaryentry{NB}
{
    name=NB,
    description={Naive Bayes}
}

\newglossaryentry{NLP}
{
    name=NLP,
    description={Natural Language Processing}
}

\newglossaryentry{ML}
{
    name=ML,
    description={Machine Learning}
}
\newglossaryentry{LSTM}
{
    name=LSTM,
    description={long short-term memory}
}
\newglossaryentry{BiRNN}
{
    name=BiRNN,
    description={bidirectional recurrent neural network}
}

\newglossaryentry{XGBoost}
{
    name=XGBoost,
    description={extreme gradient boosting}
}

\newglossaryentry{MLP}
{
    name=MLP,
    description={multilayer perceptron}
}
\newglossaryentry{VADER}
{
    name=VADER,
    description={valence  aware  dictionary  for sentiment reasoning}
}

\newglossaryentry{LR}
{
    name=LR,
    description={logistic regression}
}
\newglossaryentry{LDA}
{
    name=LDA,
    description={Latent Dirichlet Allocation}
}
\newglossaryentry{KNN}
{
    name=KNN,
    description={k-nearest neighbors}
}

\newglossaryentry{HMP}
{
    name=HMP,
    description={hostile media phenomenon}
}
\newglossaryentry{FMP}
{
    name=FMP,
    description={Friendly Media Phenomenon}
}
\newglossaryentry{CNN}
{
    name=CNN,
    description={convolutional neural network}
}

\makeglossaries

\usepackage{array}
\usepackage{ltablex}
\usepackage{longtable}
\newcolumntype{P}[1]{>{\raggedright\arraybackslash}p{#1}}
\usepackage{geometry} 
\usepackage{pdflscape}
\usepackage{graphicx}
\usepackage{cleveref}
\usepackage{longtable}
\usepackage{tablefootnote}
\usepackage{threeparttable}
\usepackage{todonotes}
\AtBeginDocument{%
  \providecommand\BibTeX{{%
    \normalfont B\kern-0.5em{\scshape i\kern-0.25em b}\kern-0.8em\TeX}}}

\setcopyright{acmcopyright}
\copyrightyear{2023}
\acmYear{2023}
\acmDOI{10.1145/1122445.1122456}

 \acmJournal{CSUR}

\begin{document}

\title[The Media Bias Taxonomy]{The Media Bias Taxonomy: A Systematic Literature Review on the Forms and Automated Detection of Media Bias} 


\author{Timo Spinde}
\affiliation{%
  \institution{University of Göttingen}
  \country{Germany}
  \city{Göttingen}
  }
  \email{Timo.Spinde@uni-konstanz.de}

\author{Smi Hinterreiter}
\authornote{Both authors contributed equally to this research.}
\affiliation{%
  \institution{University of Würzburg}
  \city{Würzburg}
  \country{Germany}
}

\author{Fabian Haak}
\authornotemark[1]
\affiliation{%
  \institution{TH Köln - University of Applied Sciences}
  \city{Köln}
  \country{Germany}
  }
  \email{fabian.haak@th-koeln.de}

\author{Terry Ruas}
\affiliation{%
  \institution{University of Göttingen}
  \country{Germany}
    \city{Göttingen}
    }
\email{ruas@uni-goettingen.de}

\author{Helge Giese}
\affiliation{%
  \institution{Charité – Universitätsmedizin Berlin}
  \city{Berlin}
  \country{Germany}
}
\email{helge.giese@uni-konstanz.de}

\author{Norman Meuschke}
\affiliation{%
  \institution{University of Göttingen}
  \country{Germany}
    \city{Göttingen}
    }
\email{Meuschke@uni-goettingen.de}

\author{Bela Gipp}
\affiliation{%
  \institution{University of Göttingen}
  \country{Germany}
    \city{Göttingen}
    }
\email{gipp@uni-goettingen.de}

\renewcommand{\shortauthors}{Spinde et al.}

\begin{abstract}
The way the media presents events can significantly affect public perception, which in turn can alter people's beliefs and views. Media bias describes a one-sided or polarizing perspective on a topic. This article summarizes the research on computational methods to detect media bias by systematically reviewing 3140 research papers published between 2019 and 2022. To structure our review and support a mutual understanding of bias across research domains, we introduce the Media Bias Taxonomy, which provides a coherent overview of the current state of research on media bias from different perspectives. 
We show that media bias detection is a highly active research field, in which transformer-based classification approaches have led to significant improvements in recent years. These improvements include higher classification accuracy and the ability to detect more fine-granular types of bias. However, we have identified a lack of interdisciplinarity in existing projects, and a need for more awareness of the various types of media bias to support methodologically thorough performance evaluations of media bias detection systems.
Concluding from our analysis, we see the integration of recent machine learning advancements with reliable and diverse bias assessment strategies from other research areas as the most promising area for future research contributions in the field.

\end{abstract}



\maketitle
\thispagestyle{preprintbox}

\section{Introduction} \label{sec:introduction}

Online news articles have become a crucial source of information, replacing traditional media like television, radio broadcasts, and print media (e.g., newspapers, magazines) \cite{dallmann2015a}.
However, news outlets often are biased \cite{wolton2017a}. The primary reason for this bias is that opinionated, entertaining, and sensationalist content is more likely to attract a larger audience while being less expensive to produce \cite{atkins2016a}.

Media bias is widely recognized as having a strong impact on the public's perception of reported topics \cite{dallmann2015a, Hamborg2019inter, lim2018b}. 
Media bias aggravates the problem known as filter bubbles or echo chambers \cite{Spinde2022a}, where readers consume only news corresponding to their beliefs, views, or personal liking \cite{lim2018b}. 
The behavior likely leads to poor awareness of particular issues, a narrow and one-sided perspective \cite{spinde2020a}, and can influence voting behavior \cite{Druckman_2005, deVreese_2005}. 

Highlighting media bias instances has positive implications and can mitigate the effects of such biases \cite{baumer2015a}. 
While completely eliminating bias may be an unrealistic goal, drawing attention to its existence by informing readers that content is biased allows them to compare content easily. 
It can also enable journalists and publishers to assess their work objectively \cite{dallmann2015a}. 
In the following, we list systems designed to help readers mitigate the effects of media bias on their decision-making.
Most of these systems focus on aggregating articles about the same event from various news sources to provide different perspectives \cite{lim2018b}. 
For example, news aggregators like Allsides\footnote{\url{https://www.allsides.com}} and Ground News\footnote{\url{https://ground.news}} allow readers to compare articles on the same topic from media outlets known to have different political views. 
Media bias charts, such as the AllSides media bias chart\footnote{\url{https://www.allsides.com/media-bias/media-bias-chart}} or the Ad Fontes media bias chart\footnote{\url{https://www.adfontesmedia.com/}} provide up-to-date information on media outlets' political slants. 
However, it is uncertain whether readers have the possibility and, more importantly, the desire to read several articles on the same topic and compare them. 
 
Media bias has become the subject of increasing interdisciplinary research, particularly in automated methods to identify bias. However, the concept of media bias remains loosely defined in the literature \cite{Hamborg2019inter}. Existing work uses different subcategories and types of bias \cite{Recasens_2013, Spinde_2021f}, but authors tend to focus on only one media bias subcategory while disregarding similar kinds of bias concepts. 
publications on media bias often work on similar concepts but assign different names to them, leading to confusion and imprecise use of terms. For example, some authors refer to word-based bias as linguistic bias \cite{Recasens_2013}, while others call it bias by word choice \cite{Spinde_2021a}, but the exact difference or overlap between these terms is undefined. 
The lack of clarity surrounding media bias can have negative effects on measuring media bias perception \cite{Spinde2021e}. Additionally, recent advances in Deep Learning have shown how awareness of tasks within complex domains, such as media bias, could potentially lead to large performance increases \cite{DBLP:journals/corr/abs-2111-10952}. However, these advancements have yet to be incorporated into media bias research \cite{spinde_exploiting_2022}. 

Our literature review seeks to create awareness of media bias detection as a task and to provide a summary of existing conceptual work on media bias and automated systems to detect it. To achieve this, we compare and contrast computer science research while also incorporating media bias-related concepts from non-technical disciplines such as framing effects \cite{entman2010media}, hate speech \cite{Davidson_2017}, and racial bias \cite{Dijk2007Chapter1D}.

We propose a unified taxonomy for the media bias domain to mitigate ambiguity around its various concepts and names in prior work. 
In addition, we classify and summarize computer science contributions to media bias detection in six categories\footnote{We reason and detail our categories in \Cref{csmethods}.}: 
(1) traditional natural language processing (\gls{tNLP}) methods \cite{niven_measuring_2020}, (2) simple non-neural \gls{ML} techniques \cite{shahid_detecting_2020}, (3) transformer-based (\gls{tbML}) \cite{ sinno_political_2022} and (4) non-transformer-based (\gls{ntbML}) \cite{fagni_fine-grained_2022} machine learning. We also include (5) non-neural network (\gls{nNN})-based (\Cref{sec:oML}) \cite{rao2021gender} as well as graph-based \cite{guimaraes_characterizing_2021} approaches. 
Lastly, we provide an overview of available datasets.
Our aim is to provide an overview of the current state-of-the-art in media bias and increase awareness of promising methods. We show how computer science methods can benefit from incorporating user and perception-related variables in different datasets to improve accuracy. To facilitate the usage of such variables, we give an overview of recent findings about cognitive processes behind media bias.
We believe that a systematic overview of the media bias domain is overdue given the numerous papers covering related issues. Such an overview can benefit future work in computer science and other areas, such as Psychology, Social Science, or Linguistics, which all cover media bias.
As we show in detail in \Cref{sec:rel_lit}, existing literature reviews on media bias \cite{Hamborg2019inter, Nakov2021ASO, anderson_chapter_2015} do not cover crucial aspects. They do not give a systematic overview of related concepts, instead presenting how media bias can develop. 
Aside from the major developments within the media bias domain since 2021, they lack details on computer science methods and psychological and social science research. 

In summary, our literature review answers the following research questions:
\begin{enumerate}
    \item[(RQ1)] What are the relationships among the various forms of bias covered in the literature? 
    \item[(RQ2)] What are the major developments in the research on automated methods to identify media bias?
    \item[(RQ3)] What are the most promising computer science methods to automatically identify media bias?
    \item[(RQ4)] How does social science research approach media bias, and how can social science and computer science research benefit each other?
\end{enumerate}

All resources for our review are publicly available at \label{taxonomyurl}\url{https://github.com/Media-Bias-Group/Media-Bias-Taxonomy}.


\section{Methodology} \label{sec:methodology}

The core contribution of this article is a systematic literature review that provides a structured and comprehensive overview of the application of computer science methods for detecting media bias.
This review also clarifies and establishes connections among the various concepts employed in the field of media bias.
Reviews are susceptible to incomplete data and deficiencies in the selection, structure, and presentation of the content \cite{Fagan_2017}, especially when aiming for extensive coverage. 
To overcome these challenges, we designed our collection and selection processes carefully, with a focus on mitigating common risks associated with literature reviews.

We used automated, keyword-based literature retrieval (described in \Cref{sec:crawl}), followed by a manual selection (\Cref{sec:review}), and adhered to established best practices for systematic literature reviews \cite{10.1145/3345317, 10.1145/3563691, 10.1145/3555719}. 

The number of concepts (and keywords) relevant to media bias is high but hard to define.\footnote{For example, the term bias also yields many health-related papers that are irrelevant to our review.}
Reviewing all papers for all related concepts is unfeasible\footnote{Based on the keywords we searched for, which we detail in \Cref{sec:crawl}, we found over 100.000 publications.}. Therefore, we applied filter criteria to select candidate documents. 
Moreover, we excluded references from the selected papers as additional candidates since determining an unbiased stopping criterion would be challenging.
Our review covers the literature published between January 2019 and May 2022, thus providing a comprehensive overview of the state-of-the-art in the field.

To ensure diversity in the computer science publications included in our review, we retrieved literature from two sources: DBLP (DataBase systems and Logic Programming)\footnote{\url{https://dblp.org/}} and Semantic Scholar\footnote{\url{https://www.semanticscholar.org/}}. 
Both sources are reliable and diverse and therefore meet the criteria for suitable sources for literature reviews \cite{Kit04, BRERETON2007571}. 
DBLP is the most extensive database for computer science publications to date, containing documents from major peer-reviewed computer science journals and proceedings. 
It is a primary literature platform used in other reviews \cite{10.1145/3433000, 10.1145/3554727, 10.1145/3453476}. 
Semantic Scholar draws on a considerably larger database than DBLP, going beyond computer science into other research areas. 
It is also frequently used in literature reviews \cite{hanousse, 10.1145/3209978.3209982, 10.1145/3404835.3462788} and allows for applying more filter criteria to searches, particularly filtering by scientific field.

Both platforms are accessible through an API and facilitate the use of an automated retrieval pipeline, which we require to filter our search results efficiently.
We retrieved results for a selection of search terms (see \Cref{sec:crawl}). 
While Semantic Scholar is an extensive general knowledge archive, DBLP focuses on in-depth coverage of computer science. 
By including both major archives, we aim to retrieve an exhaustive set of candidate documents in computer science. 

\subsection{Retrieving Candidate Documents} \label{sec:crawl} 
\begin{figure}[H] 
    \centering
    \includegraphics[width=1\textwidth]{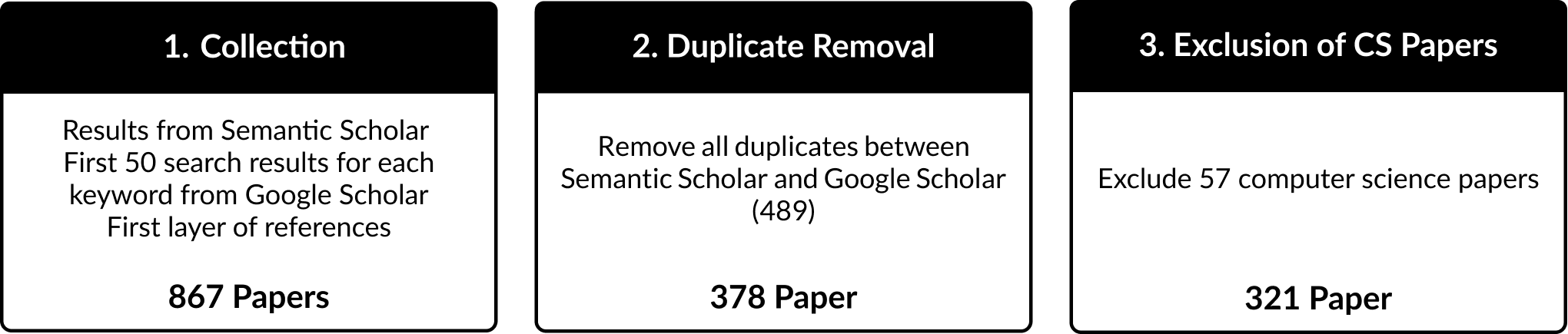}
    \caption{Number of publications at each step of the literature retrieval and review of computer science publications.}
    \label{fig:cs_crawl}
\end{figure}

We used media bias terms encountered during our initial manual retrieval step (depicted in \Cref{fig:cs_crawl}) as search queries to create candidate lists for our literature review.\footnote{Initially, we used more general terms such as media bias'', hate speech'', linguistic bias'', and racial bias'' which are widely known.
We manually identified additional bias concepts in the retrieved publications during our searches depicted in \Cref{fig:cs_crawl} and \Cref{fig:cs_crawl_tax} and added them to our list of search queries.
Subsequently, we searched for these newly identified keywords, creating the media bias keyword list presented in \Cref{fig:framework}.}  
These terms also served as the basis of the media bias categories we consolidated in our Media Bias Taxonomy in \Cref{sec:framework}.
In step 2 (\Cref{fig:cs_crawl}), we employed a Python pipeline to retrieve computer science documents from both DBLP and Semantic Scholar, merge and unify the search results, and export them as tabular data.\footnote{We have made the crawler publicly available for use in other projects.
The code and instructions can be found in our \hyperref[taxonomyurl]{repository}.} 
We scraped a list of 1496 publications from DBLP and 1274 publications from Semantic Scholar for the given time frame.
We present the complete list and search keywords in our \hyperref[taxonomyurl]{repository}.
As shown in \Cref{fig:cs_crawl}, we obtained a list of 3140 candidates for the literature review.
After removing 531 duplicates between the Semantic Scholar and DBLP results, the final list contained 2609 publications.
All search results were tagged with the relevant search queries and exported as a CSV file for the selection step.

\subsection{Candidate Selection} \label{sec:review}
We followed a  multi-stage process to select relevant publications, as shown in \Cref{fig:cs_crawl}. The figure also shows the number of publications in each step. 
Three reviewers (Ph.D. students in computer science) filtered the results after the automatic scrape (step 2) and duplicate removal (step 3).
In step 4, they filtered for documents that cover media bias, based on the title, abstract, and text, which resulted in 299 documents. 
In step 5, one reviewer per paper thoroughly inspected every publication to investigate whether computer science methods were used to detect media bias.
For each publication, we exported the used methods and datasets (see \Cref{csmethods}). 
In step 6, a second reviewer verified the choice of the first reviewer for each publication. 
In case of disagreement or uncertainty, the third reviewer was consulted.
For each publication, at least two of the three reviewers must deem the publication suitable for our review.
The detailed selection criteria for each step are available in our \hyperref[taxonomyurl]{repository}. 
In the end, we selected 96 relevant documents. We assigned each paper to its computer science methods category according to \Cref{fig:cs_overview}.

\subsection{Finding Additional Conceptual Literature for the Media Bias Taxonomy} \label{sec:review_concepts}
\begin{figure}[H] 
    \centering
    \includegraphics[width=1\textwidth]{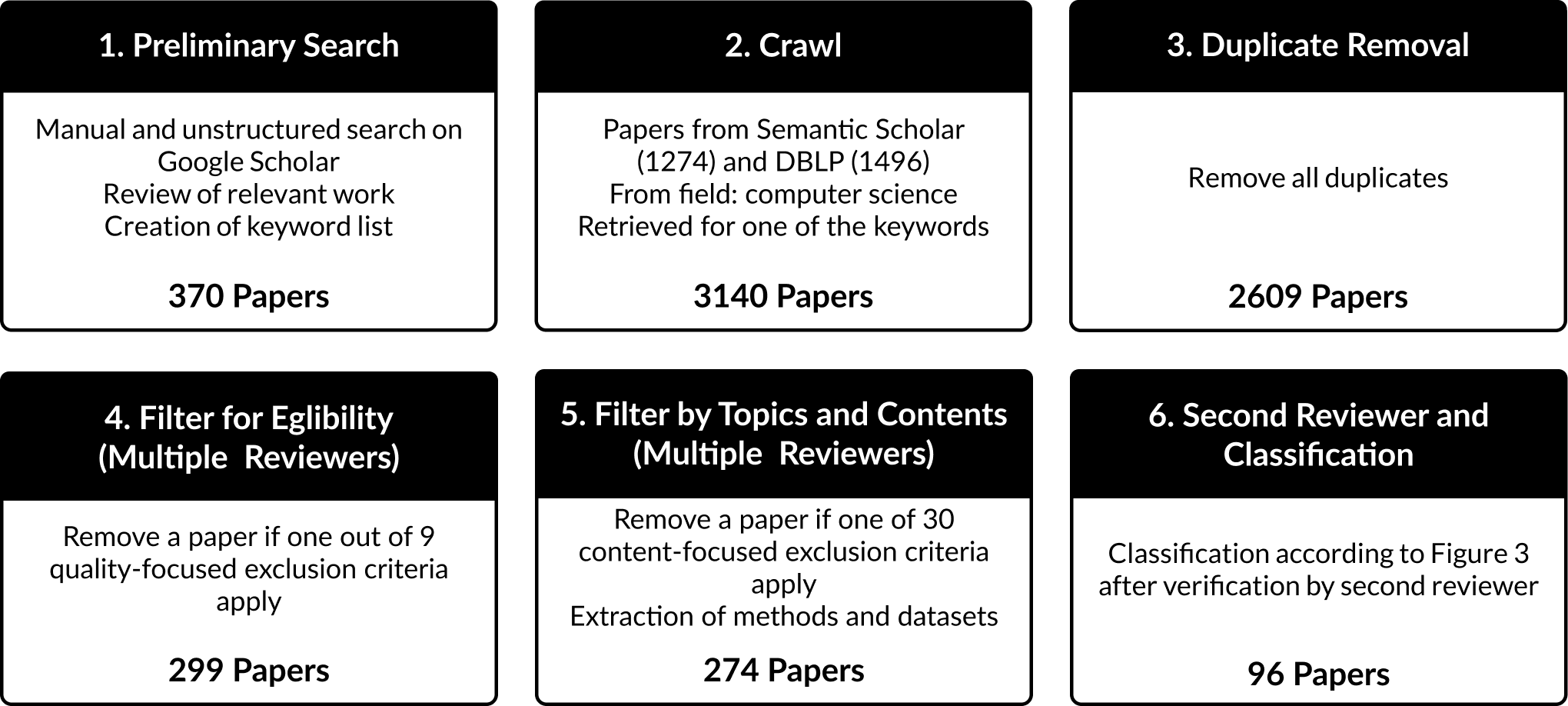}
    \caption{Number of publications at each step of the literature retrieval and review for the Media Bias Taxonomy.}
    \label{fig:cs_crawl_tax}
\end{figure}

One goal of our systematic literature review is to develop a taxonomy that organizes the various definitions of media bias into distinct types. 
However, while conducting our search, we recognized that most computer science publications focus on methodology rather than defining bias types. 
Therefore, we expanded our search to other research areas that may have different perspectives on media bias. 
For this purpose, we conducted a second search, as shown in \Cref{fig:cs_crawl_tax}, replacing DBLP with Google Scholar to identify more non-computer science research\footnote{Google Scholar is also a reliable and diverse database, meeting the criteria recommended in systematic literature review guidelines \cite{Kit04, BRERETON2007571}.}.
We manually selected papers from the first 50 search results for each keyword on Google Scholar and Semantic Scholar\footnote{In this step, we excluded computer science publications in the Semantic Scholar results.} and checked the first layer of their references for additional relevant literature.

Overall, the additional search step for non-computer-science publications yielded 867 results, of which 489 were duplicates between Google Scholar and Semantic Scholar. 
Of the 378 non-duplicate publications, 57 were included in the search for computer science publications. 
We present the results of our searches in \Cref{sec:framework}\footnote{Due to space restrictions we do not cite all of the filtered works in this article but omit publications focusing on highly similar concepts.}.

\section{Related Literature Reviews} \label{sec:rel_lit}
Related literature reviews\footnote{We considered a publication a literature review if its main focus is a critical summary and evaluation of research about a topic related to media bias.} on media bias are scarce. 
Our literature crawl and search (\Cref{sec:methodology}) yielded only three such results \cite{Hamborg2019inter, Nakov2021ASO, anderson_chapter_2015}. 
An additional search for the terms ``media bias'' and ``news bias''\footnote{We manually examined the first 50 results on Google Scholar.} on Google Scholar did not yield more findings.
In their literature review, 
\citet{Hamborg2019inter} defined sub-categories of media bias from a social science perspective and showed how they emerge during journalistic work. 
Further, the authors described the advancements in computer science and indicated that frame analysis exists in both social sciences and computer science.

In the second work, \citet{Nakov2021ASO} surveyed media profiling approaches. 
They summarized computer science methods to analyze factuality (i.e., stance and reliability) and various forms of media bias (selection bias, presentation bias, framing bias, and news slant).
The authors separated four prediction bases for media bias: 1) textual content and linguistic features, 2) multimedia content, 3) audience homophily, and 4) infrastructure characteristics.

Lastly, \citet{anderson_chapter_2015} surveyed the literature on media bias from a sociological perspective and offered an overview of possible bias measurements.
They grouped biases into three kinds of measurement: comparing media outlets with other actors, the intensity of media coverage, and tone. 

The earlier literature reviews exhibit three major shortcomings. 
First, both computer science-focused reviews \cite{Hamborg2019inter, Nakov2021ASO} lack a systematic literature search. They only covered selected computer science approaches and datasets. 
Second, \citet{Hamborg2019inter} and \citet{Nakov2021ASO} did not cover the psychological perspective on bias, which we argue is essential to create and evaluate detection methods and datasets \cite{Spinde2021e}. 
Third, no work thus far has provided a detailed overview of the various concepts and subcategories that fall under the umbrella term media bias.
Current literature on media bias often addresses related concepts like hate speech, gender bias, and cognitive bias, but uses the umbrella term of media bias without clearly differentiating between overlapping categories and their relationships.

To our knowledge, we are the first to offer a large-scale, systematic analysis of the media bias domain. 
As a result, we provide our Media Bias Taxonomy, which connects the various definitions and concepts in the area. 
In addition, we briefly summarize the state-of-the-art psychological research on media bias and provide an in-depth overview of all computer science methods currently used to tackle media bias-related issues.

Our review focuses exclusively on media bias and does not include publications on related topics such as fake news. For details on fake news and its detection, we recommend referring to the two literature reviews \cite{doi:10.1080/23808985.2019.1602782, https://doi.org/10.1111/soc4.12724}.

\section{Related Work and Theoretical Embedding} \label{sec:related}

This section will provide an overview of media bias, followed by a presentation and organization of related concepts in our novel Media Bias Taxonomy.


\subsection{Media Bias} \label{sec:bias}
Media bias is a complex concept \cite{Spinde2021e, Spinde_2020} that has been researched at least since the 1950s \cite{White_1950}. 
It describes slanted news coverage or other biased media content \cite{Hamborg2019inter}, which can be intentional, i.e., purposefully express a tendency towards a perspective, ideology, or result \cite{williams1975a}, or unintentional \cite{williams1975a, baumer2015a}. 
Different stages of the news production process can introduce various forms of media bias \cite{Hamborg2019inter}.

The lack of a precise and unified definition for media bias, sometimes referred to as editorial slant \cite{Druckman_2005}, has contributed to the conceptual fragmentation in the field \cite{Spinde2021e}. For instance, \citet{D'Alessio_2000} categorized media bias into three primary groups \cite{D'Alessio_2000}: gatekeeping bias, coverage bias, and statement bias. In contrast, \citet{Mullainathan_2002} proposed two types of media bias: ideology bias and spin bias \cite{Mullainathan_2002}. Some scholars referred to media bias as lexical or linguistic bias \cite{Beukeboom_2017}. 
Others have proposed less specific definitions. For instance, \citet[p. 2]{Spinde_2021a} described media bias as ``slanted news coverage or internal bias reflected in news articles.'' \citet[p. 1268]{Lazaridou_2020} defined it as news reporting that ``leans towards or against a certain person or opinion by making one-sided misleading or unfair judgments,'' and \citet[p. 1]{Lee_2021} defined it as reporting ``in a prejudiced manner or with a slanted viewpoint.'' None of these definitions is based on a comprehensive literature review.
Therefore, we provide a comprehensive and well-organized description of media bias in \Cref{sec:framework}, which includes its sub-fields and related computer science methods and discuss the common ground of all media bias concepts in our review in \Cref{sec:discussion}.

It is worth mentioning that media bias does not only manifest via text but also via pictures or text/news layout \cite{Matthes_2021, Peng_2018}. 
Moreover, biased reporting in one outlet can also cause biased reporting in other outlets by direct citations \cite{doi:10.1080/08913810508443641}. 
Our literature review focuses on text-based media bias and methods only. 

%
%
\subsection{The Media Bias Taxonomy} \label{sec:framework}
 
\begin{figure}[H] 
  \centering
  \includegraphics[width=\textwidth]{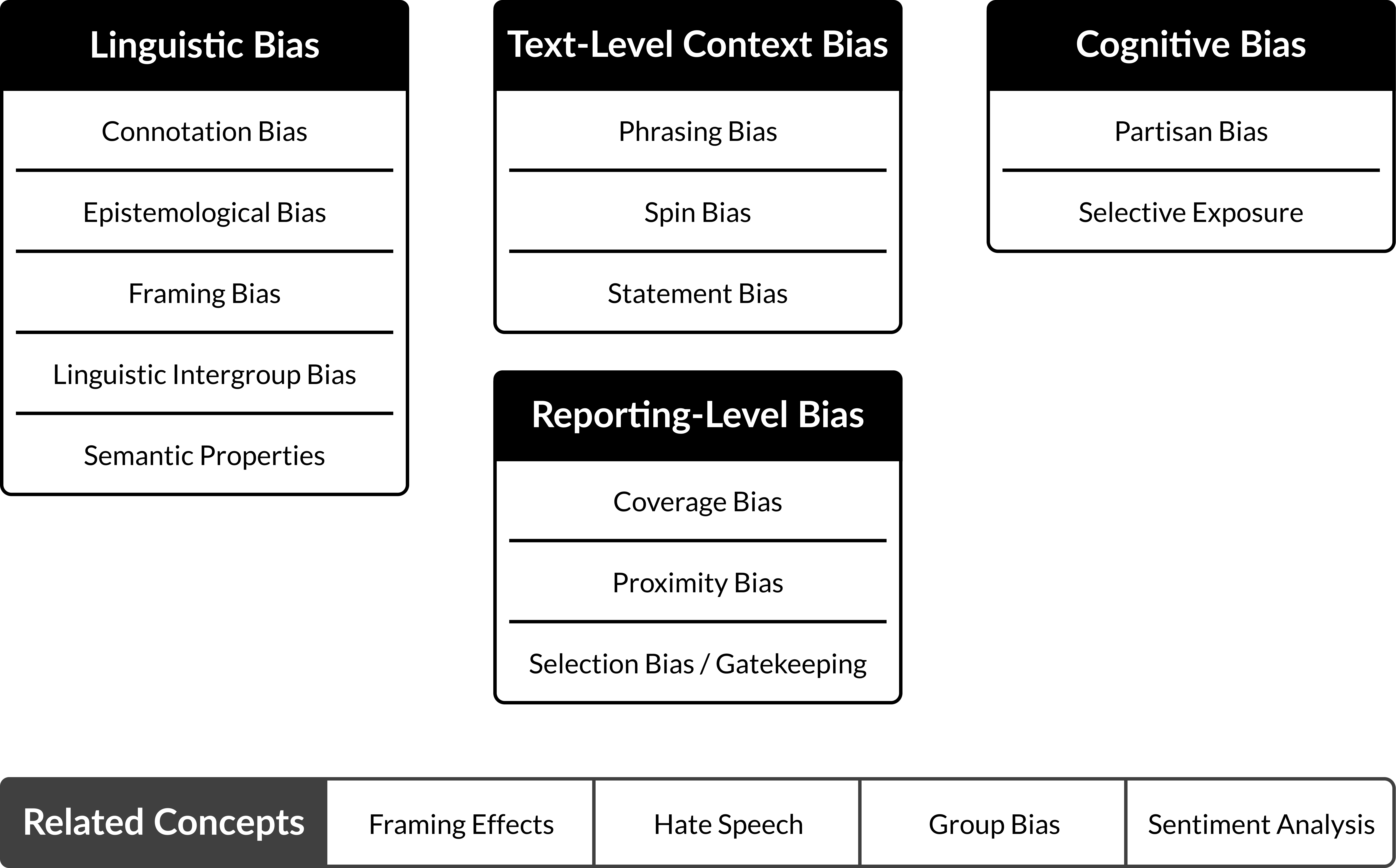}
  \caption{The Media Bias Taxonomy. The four subcategories of media bias consist of different bias types.} 
  \label{fig:framework}
\end{figure}

As media bias definitions often overlap, a clear distinction between its types is challenging. 
We propose the Media Bias Taxonomy, depicted in \Cref{fig:framework} to give a comprehensive overview of the media bias domain. 
Based on a manual selection after the literature search process, described in \Cref{sec:review_concepts}, we split media bias into four major bias categories: linguistic, cognitive, text-level context, reporting-level, as well as related concepts, which are detailed in the following subsections. 
We show detailed examples in \Cref{app:1} for all subtypes of bias\footnote{Other, overarching concepts exist, such as persuasiveness \cite{greenPersuasivenessNarratives2005}, which we do not cover or organize within this work. In future work, we will address concepts containing mmultiple forms of bias}.     

%
%


\subsubsection{Linguistic Bias} \label{sec:linguistic_bias}
Linguistic bias, sometimes called lexical bias \cite{fan_plain_2019}, refers to a pattern of using certain words that reflects a particular way of thinking about a group or an individual based on their social category. This bias involves a systematic preference for certain words or phrases that may reflect stereotypes or preconceived notions about the group or individual being described \cite{Beukeboom_2017}. In simpler terms, linguistic bias means using language that reflects a particular attitude or viewpoint towards a particular group or individual.

We identified five bias types within this category: linguistic intergroup bias \cite{Semin_1988}, framing bias \cite{Recasens_2013}, epistemological bias \cite{Recasens_2013}, bias by semantic properties \cite{Greene_2009}, and connotation bias \cite{Rashkin_2016}.
\Cref{table:categories1} lists examples for each subcategory. 

\textbf{Linguistic Intergroup Bias} describes which group members use specific language \cite{Semin_1988}. 
The concept is based on the linguistic category model (LCM), which categorizes words into different levels of abstraction (action words, interpretive action words, state verbs, and adjectives) according to their purpose \cite{Semin_1988, Dragojevic_2016}.  
The use of biased language is often subtle and reinforces stereotypes \cite{Maass_1989, Beukeboom_2017}.
\citet{Maass_1989} illustrated linguistic intergroup bias with the following example: 

\begin{itemize}
    \item They considered the hypothetical scenario where ``Person A is hitting Person B's arm with his fist'' \cite[p. 982]{Maass_1989}. 
    \item  Describing the scenario using the least abstract form of language, one could say, ``A is punching B'' \cite[p. 982]{Maass_1989}. This entails no kind of valuation or implication and only describes what happened.
    \item  In contrast, using the most abstract form of language, one could say ``A is aggressive'' \cite[p. 982]{Maass_1989}. This might or might not be accurate and cannot be judged from the fact that A hit B.
\end{itemize}

\textbf{Framing Bias} is defined as the use of ``subjective words or phrases linked with a particular point of view'' \cite[p. 1650]{Recasens_2013} to sway the meaning of a statement. 
The subjective words are often either one-sided terms or subjective intensifiers \cite{Recasens_2013}. 
One-sided terms are words that ``reflect only one of the sides of a contentious issue'' \cite[p. 1653]{Recasens_2013}, while subjective intensifiers are adjectives or adverbs that reinforce the meaning of a sentence. 

\textbf{Epistemological Bias} describes the use of linguistic features that subtly focus on the credibility of a statement \cite{Recasens_2013}.
Word classes associated with epistemological bias are factive verbs, entailments, assertive verbs, and hedges, see examples in \Cref{table:categories1}. 
Factive verbs indicate truthfulness; entailments are relations where one word implies the truth of another word. 
Assertive verbs state clearly and definitely that something is true.
Hedges are words used to introduce vagueness to a statement. 
In contrast to framing bias, epistemological bias is rather subtle and implicit \cite{Recasens_2013}. 

\textbf{Bias by Semantic Properties} describes how word choice affects the framing of content and triggers bias, similar to framing bias and epistemological bias. 
The difference, however, is that framing and epistemological bias refer to the individual words used, whereas bias by semantic properties refers to how the sentence is structured \cite{Greene_2009}. 

\textbf{Connotation Bias} refers to using connotations to introduce bias to a statement  \cite{Rashkin_2016}. 
While the denotation of a word expresses its literal meaning, the connotation refers to a secondary meaning besides the denotation. 
The connotation is usually linked to certain feelings or emotions associated with a point of view \cite{Rashkin_2016}.
%
%
\subsubsection{Text-level Context Bias} \label{sec:text_level_bias}
Similar to linguistic bias, text-level context bias refers to the way the context of a text is expressed. 
Words and statements have the power to alter the article's context, influencing the reader's opinion \cite{Hube_2018b}. 
The types of bias belonging to this category are statement bias \cite{D'Alessio_2000}, phrasing bias \cite{Hube_2018b}, and spin bias \cite{Alonso_2017}, which consists of omission bias and informational bias \cite{Alonso_2017}. \Cref{table:categories2} lists examples for each subcategory. 

\textbf{Statement Bias} refers to ``members of the media interjecting their own opinions into the text'' \cite[p. 136]{D'Alessio_2000}, which leads to certain news being reported in a way that is more or less favorable towards a particular position \cite{D'Alessio_2000}. 
These opinions can be very faint and are expressed ``by disproportionately criticizing one side'' \cite[p. 250]{Budak_2016} rather than ``directly advocating for a preferred [side]'' \cite[p. 250]{Budak_2016}. 

\textbf{Phrasing Bias} is characterized by inflammatory words, i.e., non-neutral language \cite{Hube_2018b}. 
Depending on the context, a word can change from neutral to inflammatory. 
Therefore, when analyzing bias, the inter-dependencies between words and phrases must be considered \cite{Hube_2018b}. 

\textbf{Spin Bias} describes a form of bias introduced either by leaving out necessary information \cite{Mullainathan_2002, Alonso_2017} or by adding unnecessary information \cite{fan_plain_2019}. 
The underlying motivation is to tell a simple and memorable story \cite{Mullainathan_2002}.
Spin bias can be divided into omission, and informational bias \cite{Alonso_2017}. 
Omission bias, also known as simplification, is the act of omitting words from a sentence \cite{Alonso_2017, Mullainathan_2002}.
Informational bias, or exaggeration, is defined as adding speculative, tangential, or irrelevant information to a news story \cite{fan_plain_2019}.
%
%

\subsubsection{Reporting-level Context Bias} \label{sec:reporting_level_bias}
Reporting-level context bias subsumes all bias types on the reporting level. 
While text-level context bias observes bias within an article, reporting-level bias observes the general attention for specific topics \cite{Galtung_1965, D'Alessio_2000, Saez-Trumper_2013, Budak_2016}. 
Bias types in this category are selection bias, proximity bias, and coverage bias, which are all closely connected. 
\Cref{table:categories3} lists examples for each subcategory. 

\textbf{Selection Bias} (or gatekeeping bias) refers to the selection of content from the body of potential stories by writers and editors \cite{D'Alessio_2000}. 
Obviously, not all news events can be reported due to the limited resources of newspapers. 
However, this decision-making process is prone to bias from personal preferences \cite{White_1950, D'Alessio_2000, Saez-Trumper_2013, Mullainathan_2002}. 

\textbf{Coverage Bias} describes situations in which two or more sides of an issue receive imbalanced amounts of attention, such as pro-life vs. pro-choice statements \cite{D'Alessio_2000}.\footnote{Coverage bias refers to a particular event, whereas reporting-level context bias refers to the general attention a topic receives.}
The level of attention can be measured either in absolute numbers (e.g., there are more articles discussing pro-life than pro-choice topics), how much space the topics get in a newspaper (e.g., printed on the front page), or as the length of the article (e.g., pro-life articles are longer and receive more in-depth coverage than pro-choice articles) \cite{Saez-Trumper_2013, D'Alessio_2000}. 

\textbf{Proximity Bias} focuses on cultural similarity and geographic proximity as decisive factors. 
Newspapers tend to report more frequently and more in-depth on events that happened nearby \cite{Saez-Trumper_2013}. 
For instance, the more two countries are culturally similar, the more likely it is that events from one region or country will be reported in the other, and the coverage will be more in-depth \cite{Saez-Trumper_2013, Galtung_1965}. 
%
%


\subsubsection{Cognitive Bias} \label{sec:cognitive_bias}
The processing of media information may also be biased by the reader of an article and the state the reader is in during reading. 
In this review, we use the term cognitive bias, defined as ``a systematic deviation from rationality in judgment or decision-making'' \cite[p. 1]{Blanco_2017}, to summarize how this processing may be negatively affected. 
While a failure to detect biased media in a given set of articles may be explained by a lack of ability or motivation (e.g., being inattentive/ disinterested, focusing on identity instead of accuracy motives), biased processing of news by the reader is often attributed to a need for a consistent world view and for overcoming dissonances evoked by discordant information \citep{10.1037/1089-2680.2.2.175}. 
In this line of reasoning, repeated exposure and increased familiarity with an argument as well as source cues for a reputable, world-view-consistent source, may increase the trust in information quality.

\textbf{Selective Exposure.} 
Similar to the selection bias of editors and authors, readers also actively select which articles they read \cite{klapper1960effects}. 
Given this choice, they tend to favor reading information consistent with their views, exacerbating already existent biases through selective exposure to one-sided news reports \cite{10.1002/asi.24121, 10.1073/pnas.1517441113}. 
Additionally, such selective exposure tends to extend to social tie formation. Topic information is solely exchanged among like-minded individuals, a phenomenon often dubbed echo chamber or filter bubble \cite{10.1371/journal.pone.0147617}\footnote{In case an algorithm was trained to this preference.}, hampering unbiased information processing.

\textbf{Partisan Bias.} 
Selective attention to world-view-consistent news has led to research on the effects of political identity. There, the evaluation of veracity seems dependent on the fit to the reader's party affiliation, a phenomenon dubbed partisan bias \cite{10.1016/j.tics.2021.05.001, 10.1177/1745691620986135}. Similarly, the hostile media phenomenon (HMP) describes the general observation that members of opposing groups rate a news article as biased against their point of view \cite{Vallone_1985}.

\subsubsection{Related Concepts} \label{sec:related_concepts}
The last category contains definitions that cannot be exclusively assigned to any other media bias category. 
Concepts belonging to this category are framing effects \cite{deVreese_2005}, hate speech \cite{Davidson_2017}, sentiment analysis, and group bias \cite{chiazor2021automated}, which consists of gender bias \cite{Costa_2019}, and religion bias \cite{manzini-etal-2019-black}. 
Much research focuses on these concepts, so we introduce them only briefly and refer to other sources for more information. 

\textbf{Framing Effects} refer to how media discourse is structured into interpretive packages that give meaning to an issue, so-called frames. Frames promote a specific interpretation of the content or highlight certain aspects while overlooking others. In other words, this type subsumes biases resulting from how events and entities are framed in a text \cite{deVreese_2005, Entman_2007}.

\textbf{Hate Speech} is defined as any language expressing hatred towards a targeted group or intended to be derogatory, humiliate, or insult \cite{Davidson_2017}. 
Often, hateful language is biased \cite{Mozafari_2020}.
The consequences of hate speech in media content are severe, as it reinforces tension between all actors involved \cite{Mozafari_2020, Ali_2021}. 

\textbf{Group Bias.} 
We categorize gender bias, racial bias, and religion bias under the umbrella term ``group bias,`` as they all refer to biased views toward certain groups.

\textbf{Gender Bias} is characterized by the dominance of one gender over others in any medium \cite{Costa_2019}, resulting in the under-representation of the less dominant gender and the formation of stereotypes \cite{Costa_2019, power2019women}. It is associated with selection bias \cite{asr2021gaptracker, johannsdottir2015gender}, coverage bias \cite{bystrom2001framing, leavy2018uncovering}, and context bias at the text level. For instance, women are quoted more frequently than men for ``Lifestyle'' or ``Healthcare'' topics, while men are quoted more frequently in ``Business'' or ``Politics'' \cite{rao2021gender}. Linguistic research on gender bias aims to identify gender-specific and gender-neutral words \cite{dacon_does_2021} and create lexicons of verbs and adjectives based on gender stereotypes \cite{fast2016shirtless}.

\textbf{Racial Bias} and \textbf{Religion Bias} are other types of group bias. Racial bias refers to the systematic disproportionate representation of ethical groups, often minorities \cite{chiazor2021automated}, in a specific context \cite{Min_2010, chiazor2021automated}.

Religion, racial, and gender biases can be observed in word embeddings. For example, ``Muslim'' is spatially close to ``terrorist'' in some embeddings \cite{manzini-etal-2019-black}, which may result from biased texts in the data used to derive these embeddings (as word embeddings depend on their input).

Group biases can manifest in other forms, such as hate speech, which is a subgroup of biases. Although the distinction between racial and gender biases is not always evident, they can exist independently \cite{Min_2010, Gershon_2012}.

\textbf{Sentiment Analysis} involves examining text for its emotional content or polarity \cite{Enevoldsen_2017}. In the context of media bias, sentiment analysis can detect biases in statements or articles \cite{Hamborg_2021, Hube_2018a} and help identify other concepts like hate speech, political ideology, or linguistic bias \cite{Rodriguez_2019, Ali_2021}.
%
%

\newcolumntype{L}[1]{>{\raggedright\arraybackslash}p{#1}}
 
\newcolumntype{C}[1]{>{\centering\arraybackslash}p{#1}}
 
\newcolumntype{R}[1]{>{\raggedleft\arraybackslash}p{#1}}

\section{Computer Science Research on Media Bias} \label{csmethods}

Computer science research on media bias primarily focuses on methods used to analyze, mitigate, and eliminate bias in texts. Detecting bias is a prerequisite for other applications~\cite{raza_dbias_2022}. Bias detection systems could also be employed to check computer-generated texts for bias. Hereafter, we provide a comprehensive overview of computer science methods used in media bias research in recent years based on a systematic literature review. The methodology of the review is described in \Cref{sec:methodology}. A systematic overview of computer science methods is essential for capturing the state of media bias research and identifying research trends and gaps. To the best of our knowledge, this is the most comprehensive survey on media bias detection methods so far, as discussed in \Cref{sec:rel_lit}.

\begin{figure}[H]
    \centering
    \includegraphics[width=0.9\textwidth]{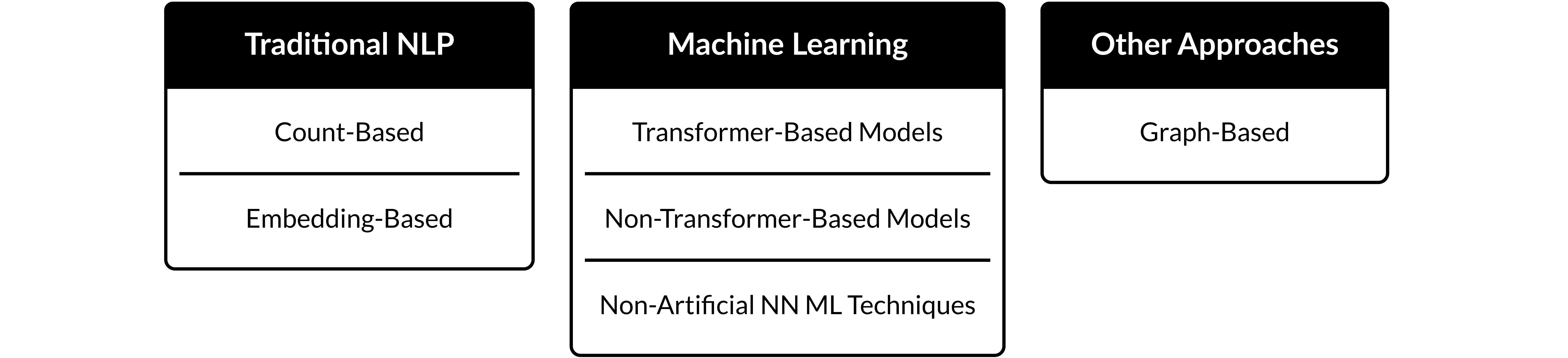}
    \caption{Classification of computer science methods for media bias detection we use in our analysis.}
    \label{fig:cs_overview}
\footnotesize
\end{figure}

\Cref{table:csmethods1} organizes the findings of our literature review by the year of publication and category of employed computer science method.\footnote{We do not report performance measures for most models, as most approaches work on different datasets and tasks, causing the scores to be incomparable. Instead, we summarize our findings on the most promising approaches at the end of this section.} We chose the employed methods as the main categorical property to structure the publications since the methods are typically described in more detail than the type of investigated bias. Our analysis shows that media bias detection methods use approaches ranging from traditional natural language processing (\gls{tNLP}) methods (e.g., \cite{niven_measuring_2020}) and simple \gls{ML} techniques (e.g., \cite{shahid_detecting_2020}) to complex computer science frameworks that combine different advanced classification approaches (e.g., \cite{Hamborg_2021}), and graph-learning-based approaches (e.g., \cite{hofmann_modeling_2021}). Therefore, we introduce the classification depicted in \Cref{fig:cs_overview}.


Approaches we classify as \gls{tNLP} (\Cref{sec:tNLP}) do not use complex \gls{ML} techniques and are commonly employed in social sciences (e.g.,~\cite{mendelsohn_framework_2020, kroon_guilty_2021}). We categorize the \gls{tNLP} publications into two groups: first, count-based techniques supported by lexical resources, and second, more sophisticated embedding-based techniques.

\gls{ML}-based approaches (\Cref{sec:MachineLearning}) are organized into transformer-based machine learning (\gls{tbML}), non-transformer-based (\gls{ntbML}), and non-neural network (\gls{nNN})-based (\Cref{sec:oML}) approaches, ordered by the frequency of application in the reviewed literature. Graph-based models represent the third major category presented in \Cref{sec:Graph-based}.

\Cref{app:2} shows the number of publications per year and category according to our search criteria (cf. \Cref{sec:methodology}). An increasing majority of publications use \gls{tbML} approaches, while the numbers of \gls{nNN}- and \gls{ntbML}-based approaches decrease. Although our review does not fully cover 2022, the numbers suggest that these trends continue.

\subsection{Traditional Natural Language Processing Techniques}\label{sec:tNLP}

The \gls{tNLP} category encompasses all publications that identify media bias using techniques not based on \gls{ML} or graph-based approaches. We include the term ``traditional'' in the category name to differentiate it from \gls{ML} and similar techniques. Moreover, techniques similar to what we label as \gls{tNLP} have already been employed in computational linguistics as early as the sixties and seventies~\cite{coling-1965-coling}. Frequently, \gls{tNLP} methods are used as a baseline when introducing new datasets due to their explainability and proven effectiveness (e.g. \cite{sales_assessing_2021, leavy_uncovering_2020, dacon_does_2021}). Furthermore, social sciences are increasingly adopting them because of their accessibility and ease of use \cite{kroon_guilty_2021}. Although some approaches leverage \gls{ML} techniques (e.g., \cite{dalonzo_machine-learning_2021}), we classify them as \gls{tNLP} if the main contribution is a non-\gls{ML} approach. The \gls{tNLP} methods can be divided into count-based and embedding-based approaches. Count-based approaches quantify words and n-grams in the text to analyze bias, while embedding-based approaches are more sophisticated and serve to represent texts for either facilitating comparisons (e.g., \cite{leavy_uncovering_2020, papakyriakopoulos_bias_2020}) or analyzing text associations and inherent biases (e.g., \cite{caliskan_semantics_2017, ferrer_discovering_2020}).

\begin{table}[H]
\begin{threeparttable}
\begin{tabular}{l|L{2.5cm}|L{2.5cm}|L{2.5cm}|L{2.5cm}|l}
 & 2019 & 2020 & 2021 & 2022 & total \\\hline
 \gls{tNLP} &\cite{wevers_using_2019, badjatiya_stereotypical_2019, cornelissen_computational_2019},\cite{sapiro-gheiler_examining_2019}\tnote{*} &\cite{zahid_modeling_2020, spinde2020b, niven_measuring_2020, ferrer_discovering_2020, papakyriakopoulos_bias_2020, de_arruda_analysing_2020, aires_information_2020, mendelsohn_framework_2020}, \cite{ leavy_uncovering_2020}\tnote{*} &\cite{spinde2020a, dalonzo_machine-learning_2021, spinde_identification_2021, kroon_guilty_2021, Kwak_2021, sales_assessing_2021, dacon_does_2021},\cite{cuellar-hidalgo_luc_2021}\tnote{*} &  & 21 \\\hline
\gls{tbML} &\cite{han_fallacy_2019, Gao_2019, dinkov_predicting_2019, fan_plain_2019, pryzant_automatically_2019} &\cite{baly_what_2020, li_using_2021, pant_towards_2020, PoliBert20, bhardwaj_investigating_2021, tran_how_2020,  Mozafari_2020, van_den_berg_context_2020, kwak_systematic_2020},\cite{baly_we_2020, mokhberian_moral_2020}\tnote{*} 
&\cite{ huguet_cabot_us_2021, blackledge_transforming_2021, kameswari_towards_2021, madanagopal_towards_2021, Hamborg_2021, Spinde2021f, sinha_determining_2021, he_detect_2021,  spinde_interdisciplinary_2021, liu_transformer-based_2021},\cite{sinno_political_2022, zhou_hidden_2021}\tnote{*} &\cite{liu_politics_2022, lee_neus_2022, spinde_exploiting_2022, raza_dbias_2022, krieger_2022} & 33 \\\hline
\gls{ntbML} &\cite{Hube_2018b, bordia_identifying_2019, ananya_genderquant_2019, gangula_detecting_2019, jiang-etal-2019-team} &\cite{field_unsupervised_2020, xiao_timme_2020, cruz_document_2020, hube_methods_2020, mathew_hatexplain_2020, jiang_comparing_2020, webson_undocumented_2020, chen_analyzing_2020} &\cite{ ruan_pseudo-labelling_2021, gerald_ki_wei_huang_hyperpartisan_2021, ruan_bias_2021, krommyda_experimental_2021, bauer_analyzing_2021} &\cite{ liu_political_2021, fagni_fine-grained_2022} & 20 \\\hline

\gls{nNN} &\cite{anthonio_team_2019, baly_multi-task_2019} &\cite{bourgeois_selection_2018,   alzhrani_ideology_2020, kameswari_enhancing_2020, ganguly_empirical_2020, shahid_detecting_2020, Lazaridou_2020} ,\cite{chen_detecting_2020}\tnote{*} &\cite{Spinde_2021f, rao2021gender, Spinde_2021a}, \cite{ orbach_out_2020}\tnote{*} &  & 13 \\\hline
graph-based &\cite{li-goldwasser-2019-encoding} &\cite{ stefanov_predicting_2020, chen_neutral_2021, chitra_analyzing_2020} &\cite{ zhou_uncovering_2021, hofmann_modeling_2021, villa_echo_2021, guimaraes_characterizing_2021} &\cite{ guo_modeling_2022} & 9  \\\hline
total & 17 & 38 & 33 & 8 & 96
\end{tabular}
\caption{Results of the literature study on computer science methods used for media bias detection.} 
\label{table:csmethods1}
\begin{tablenotes}
        \item[*] We refer to this paper in multiple sections. If a publication covers multiple categories, we assign the most used category. If two categories apply equally, we assign one based on the method performing best. 
    \end{tablenotes}
\end{threeparttable}
\end{table}

\subsubsection{Count-Based Approaches}

While recent applications of \gls{tNLP} techniques primarily employ embedding-based methods, simpler count-based approaches are still in use. Count-based approaches most commonly use word counts and a lexicon as a reference to quantify linguistic characteristics and compare texts.

\citet{niven_measuring_2020} measured the alignment of texts to authoritarian state media using a count-based methodology that leveraged the LIWC lexicon~\cite{pennebaker01} for topical categorization. Similarly, \citet{spinde2020b} applied various count-based techniques to a custom dataset of German news articles and assessed their effectiveness for media bias detection. They reported precision, recall, and $F_{1}$ scores for bias and sentiment lexicons, word embeddings, and general TF-IDF measures, evaluating the identification of human-annotated bias in their dataset. A custom bias lexicon yielded the best performance with a low $F_{1}$ score of 0.31.

\citet{sapiro-gheiler_examining_2019} employed Naive Bayes (\gls{NB}) decision tree, support vector machine (\gls{SVM}), and lasso-penalty regression models based on bag-of-word representations to classify politicians' ideological positions and trustworthiness. \citet{de_arruda_analysing_2020} used a count-based approach within an outlier detection framework to identify selection, statement, and coverage bias in political news. \citet{dalonzo_machine-learning_2021} presented a singular value decomposition (SVD) approach that predicts the newspaper that published an article based on word and n-gram frequencies. Discriminative words and n-grams were derived from a multi-stage (automatic and manual) purging process. The system generates a conditional probability distribution that enables the projection of newspapers and phrases into a left-right bias space.

\citet{zahid_modeling_2020} used a contingency table showing mention counts and polarity rates for sources (S) and entities (E) within news-related content on Twitter to calculate media bias measures based on definitions for absolute and relative media bias~\cite{10.1145/2505515.2505623}. They investigated coverage, selection, and statement bias towards specific topics and entities, and further quantified and compared the number of positive and negative reports from media outlets on Twitter.

\citet{cuellar-hidalgo_luc_2021} presented their contribution to the ICON2021 Shared Task on Multilingual Gender Biased and Communal Language Identification~\cite{kumar-etal-2021-comma}, where the goal is to classify texts as aggressive, gender biased, or communally charged. They used k-nearest neighbors (\gls{KNN}) and a mixed approach consisting of \gls{NB}, \gls{SVM}, random forest (\gls{RF}), GBM, Adaboost, and a multi-layer perceptron, for classifying texts.\footnote{This work employed both \gls{tNLP} and \gls{NN} based methods. However, since the majority of the techniques fall into the \gls{tNLP} category, we discuss it here.}

\citet{dacon_does_2021} presented a study on gender bias in news abstracts using centering resonance analysis based on specifically filtered attribute words. This technique employs rich linguistic features and graph-based techniques.

\subsubsection{Word Embedding-Based Techniques}

A second group of \gls{tNLP} techniques detects media bias by deriving word associations through word embeddings. We exclude publications that investigate bias in pre-trained word embeddings, e.g.,  to understand potential biases in systems that use the embeddings, as this analysis does not represent a media bias investigation. However, we include work that uses word embeddings as proxies to help understand biases in texts used for training the embeddings. This is typically done by constructing word embeddings based on a collection of texts and investigating associations in these embeddings (e.g., \cite{papakyriakopoulos_bias_2020, leavy_uncovering_2020, ferrer_discovering_2020}). We differentiate between sparse and dense embedding-based techniques. Sparse embeddings, primarily based on TF-IDFs, are mostly used to survey the occurrence of certain words \cite{leavy_uncovering_2020}. Dense embeddings are employed to examine associations with specific terms~\cite{wevers_using_2019, ferrer_discovering_2020}.

\textbf{Sparse Word Embeddings.}
\citet{leavy_uncovering_2020} investigated gender bias in Irish newspapers, examining various discriminative features such as TF-IDFs. Alongside \gls{ML} techniques, she used count-based \gls{tNLP} approaches to detect coverage bias towards female politicians. Employing a bag-of-words approach, TF-IDFs, and linguistic labels on word forms, she provided data for classification models and directly detected bias. For instance, she found articles mentioning spouses of female politicians four times more often than male politicians.

\textbf{Dense Word Embeddings.}
Most word embedding-based techniques in this section use methods similar to the word embedding association test (WEAT) introduced by \citet{caliskan_semantics_2017}. WEAT investigates bias in the resulting word embeddings trained on a specific text corpus by measuring the cosine similarity between two sets of tokens (e.g., male and female pronouns) and another two sets of tokens, typically topic or stereotype-based words.

\citet{ferrer_discovering_2020} explored various aspects of linguistic and gender bias on Reddit using a technique akin to WEAT, while also examining biases through count-based approaches and sentiment analysis. \citet{badjatiya_stereotypical_2019} proposed a debiasing strategy using bias-sensitive words as reference, primarily focusing on replacing bias-sensitive words with less sensitive synonyms to debias text datasets. They identified replacement words using word embeddings with different algorithms such as \gls{KNN} or a centroid function.

\citet{mendelsohn_framework_2020} primarily employed embedding-based \gls{tNLP} techniques to investigate the development of dehumanization towards the LGBTQ community in New York Times articles from 1986 to 2015. \citet{wevers_using_2019} conducted a study on gender bias in Dutch newspapers between 1950 and 1990, measuring the distance of ``three sets of target words''~\cite[p. 3]{wevers_using_2019} to two gender-representative vectors. These vectors were constructed from the average of lists of ``gender words''\cite{wevers_using_2019}, such as ``man,'' ``his,'' ``father,'' and similar terms for the male vector.

Similarly, \citet{papakyriakopoulos_bias_2020} used word embedding associations to compare gender bias in Wikipedia and social media texts. \citet{kroon_guilty_2021} analyzed implicit associations with word embeddings to detect racial bias, using the term ``ethnically stereotyped bias'' in their work. \citet{spinde_identification_2021} trained two word embedding models on slanted news corpora: one using left-wing news from HuffPost and another based on right-wing Breitbart news. They employed the Word2Vec Continuous Skip-gram architecture for training and subsequently applied a distance-based technique with their word embeddings to identify strongly biased words, beginning with biased seed words.

\citet{Kwak_2021} presented a distinct approach to bias detection based on word embeddings. They introduced a method for characterizing documents by identifying the most relevant semantic framing axes (``microframes'') that are overrepresented in the text. They then assessed the extent of bias and activity of a given microframe, ultimately providing a more detailed description of the documents. For instance, they might identify that the axis of ``depressing'' and ``cheerful'' is central to an article and then analyze the wording that led to this classification~\cite{Kwak_2021}.

\citet{sales_assessing_2021} employed a mix of \gls{tNLP} techniques based on word embeddings to detect subjectivity bias, utilizing methods such as lexicon translation and document similarity measures.

\subsection{Machine Learning}\label{sec:MachineLearning}

The following section includes publications that used \gls{ML} for bias detection.
We start by presenting transformer-based models (\gls{tbML}), which were most frequently applied in the reviewed literature, followed by non-transformer-based models (\gls{ntbML}), and non-neural network models (\gls{nNN}).
\Gls{tbML} increased in popularity after their introduction in 2017~\cite{Transformer2017}, as shown in \Cref{table:csmethods1} and \Cref{app:2}.
Transformers use self-attention to weigh the importance of input data and can be fine-tuned with specific datasets, saving time and resources \cite{Transformer2017}. 
Their universal architecture captures dependencies across domains but can over-fit in case of limited training data \cite{linSurveyTransformers2021}. 


\subsubsection{Transformer-Based Models} \label{sec:TransformerBasedModels}
Researchers frequently used \gls{tbML} to detect linguistic bias or political stance with an encoder-only architecture and bias-specific pre-training.
Most often they used BERT or models derived from it, e.g, RoBERTa~\cite{Spinde_2021f, madanagopal_towards_2021, huguet_cabot_us_2021, zhou_hidden_2021, raza_dbias_2022, van_den_berg_context_2020, krieger_2022}, DistilBERT~\cite{blackledge_transforming_2021, spinde_exploiting_2022, raza_dbias_2022}, or ALBERT~\cite{kameswari_towards_2021}.
Several papers compare the performance of BERT-based models with other transformer models, e.g. T5~\cite{krieger_2022}, BART~\cite{lee_neus_2022, he_detect_2021}, ELECTRA~\cite{Spinde2021f} or XLNet~\cite{liu_transformer-based_2021}.
BERT-based models were also applied to detect media bias in languages other than English, such as Korean ((Kor)BERT)~\cite{han_fallacy_2019}, Indian (IndicBERT)~\cite{kameswari_towards_2021} or fine-tuning BERT on African American~\cite{Mozafari_2020}.
When researchers used an encoder-decoder architecture model like BART, they used the encoder only for the detection task, while the decoder performed the debiasing task~\cite{lee_neus_2022, he_detect_2021}.
BERT-based models often outperformed other transformers for most of the tasks and groups we defined for linguistic bias~\cite{Spinde2021f, huguet_cabot_us_2021, sinha_determining_2021, raza_dbias_2022}, and for political stance detection~\cite{sinno_political_2022, dinkov_predicting_2019}, which typically associates linguistic bias with specific political stances~\cite{sinno_political_2022}.

The prevalent approach in \gls{tbML} is to create or select bias-specific datasets, fine-tune the most popular models on them, and test the performance of the encoder-only architecture by comparing $F_{1}$-scores to baselines of \gls{tNLP} methods (e.g.,~\cite{Spinde2021f, Hamborg_2021, pant_towards_2020, Gao_2019, krieger_2022}).
To facilitate the evaluation of using different transformers for identifying various media bias types, we structure our review of \gls{tbML} by the type of bias used in fine-tuning.

\textbf{Linguistic Bias.} 
Most \gls{tbML} applications focus on detecting linguistic bias.
\citet{Spinde2021f} detected bias by word choice following a distant supervision approach with BERT. 
Based on the BABE dataset, BERT outperformed RoBERTa and other \gls{ML} classifiers in their application.
In contrast, \citet{huguet_cabot_us_2021} achieved the best performance on their Us vs. Them dataset with RoBERTa.
\citet{sinha_determining_2021} also fine-tuned BERT with a custom dataset and contextual embeddings.
In addition, they parsed sentences using a \gls{GCN} model with an additional layer of bidirectional long short-term memory (\gls{LSTM}) to exploit structural information.
\citet{raza_dbias_2022} proposed a four-phase pipeline consisting of detection (DistilBERT), recognition (RoBERTa), bias masking, and debiasing.
The system, fine-tuned on the MBIC dataset~\cite{Spinde_2021f}, detected biased words, masked them, and suggested a set of sentences with new words that are bias-free or less biased.
\citet{pryzant_automatically_2019} detected and automatically transformed inappropriate subjective texts into a more neutral version.
Using a corpus of sentence pairs from Wikipedia edits, their system used BERT as an encoder to identify subjective words as part of the generation process.

\textbf{Political Stance Detection.}\label{tbML_pol_stance_det}
The second most researched classification problem is political stance detection, an umbrella term closely related to partisan bias (cf. \Cref{sec:framework}) that identifies linguistic biases to identify the political biases of authors.
\citet{sinno_political_2022} studied the ideology of specific policies under discussion and presented the first diachronic dataset of news articles annotated at the paragraph level by trained political scientists and linguists.
Their fine-tuned BERT model performed best.
\citet{dinkov_predicting_2019} integrated audio, video, metadata, and subtitles in their multimodal dataset.
In addition to the text analysis with BERT, their application included metadata and audio data through open SMILE\footnote{ \url{https://www.audeering.com/de/research/opensmile/}}, resulting in the highest accuracy.
\citet{sinno_political_2022} presented a manually annotated dataset focusing on linguistic bias in news articles. 
Based on their dataset, in addition to several BERT-based classification approaches, they used a 2-layer bidirectional \gls{LSTM} for ideology prediction, which was outperformed by all transformer-based systems.

\textbf{Framing Bias.} 
\citet{mokhberian_moral_2020} used BERT with tweet embeddings, fine-tuned on the All The News dataset\footnote{\url{https://www.kaggle.com/datasets/snapcrack/all-the-news}}, and an intensity score for moral frames classification based on the moral foundation theory\footnote{Moral foundation theory explains moral differences across cultures. For more information, see the original work by \citet{Haidt2004-HAIIEH}.}.
\citet{kwak_systematic_2020} proposed a similar BERT-based method for conducting sociological frame analysis to detect framing bias.
\citet{lee_neus_2022} proposed a system for framing bias detection and neutral summary generation from multiple news headlines of varying political leanings to facilitate balanced and unbiased news reading. They performed multi-document summarization, multi-task learning with two tasks, and based their work on BART.

\textbf{Spin/Informational Bias.}
\citet{fan_plain_2019} investigated lexical and informational bias with BERT on their BASIL dataset, which others also used in their research~\cite{van_den_berg_context_2020, liu_politics_2022, sinha_determining_2021}.
\citet{van_den_berg_context_2020} fine-tuned RoBERTa as a context-inclusive model, exploring neighboring sentences, the full article, articles on the same event from other news publishers, and articles from the same domain.
Their model is domain-and-task-adapted for informational bias detection on the BASIL corpus.
They reported that integrating event context improved classification performance.

\textbf{Racial/Group Bias.}
For group bias detection, \citet{he_detect_2021} presented DEPEN, which employs a fine-tuned BERT model to detect biased writing styles.
Subsequently, they used BART to debias and rewrite these detected sentences.

\textbf{Sentiment Analysis.}
We exclude general sentiment analysis but include publications that leveraged sentiment analysis for linguistic bias detection as a stand-in for political stance detection (cf. \Cref{tbML_pol_stance_det}).
\citet{huguet_cabot_us_2021} investigated populist mindsets, social groups, and related typical emotions using RoBERTa fine-tuned on their populist attitude dataset \textit{Us vs. Them}.
\citet{Gao_2019} utilized BERT in aspect-level sentiment classification, achieving promising performances on three public sentiment datasets\footnote{The datasets include restaurant and laptop reviews, and tweets~\cite{Gao_2019}.}. They showed that incorporating target information is crucial for BERT's performance improvement.
\citet{Hamborg_2021} applied target-dependent sentiment classification (TSC) with BERT, RoBERTa, XLNET, and a BiGRU. 
They proposed a classifier, GRU-TSC, that incorporated contextual embeddings of the sentences and representations of external knowledge sources.

\textbf{Unreliable News Detection.}
\citet{zhou_hidden_2021} used RoBERTa to detect unreliable news---a task that overlaps with media bias detection.
Further, they proposed ways to minimize selection bias when creating datasets by including a simple model as a difficulty/bias probe. 
They also suggested that future model development uses a clean non-overlapping site and date split~\cite{zhou_hidden_2021}.

\subsubsection{Non-Transformer-Based Models}\label{sec:ntbML}
This section presents publications that use non-transformer-based machine learning for media bias detection, categorized by the type of detected bias.
Most commonly, \gls{ntbML} methods are used to detect media bias at the document level, e.g., hyperpartisanship and political stance.
Despite the homogeneity of detected biases, publications using \gls{ntbML} evaluate numerous aspects of the identification methodology, including training data~\cite{sinno_political_2022, mathew_hatexplain_2020}, word embeddings~\cite{cruz_document_2020, webson_undocumented_2020, jiang-etal-2019-team}, and pseudo-labeling~\cite{ruan_bias_2021}.

\textbf{Linguistic/Text-Level Bias.}
The detection of hyperpartisanship\footnote{Hyperpartisanship is not to be confused with partisan bias as described in \Cref{sec:framework}. It describes one-sidedness that can manifest in a range of biases~\cite{kiesel-etal-2019-semeval}.} is the most common application of \gls{ntbML}.
The task's popularity is partly due to the SemEval 2019 hyperpartisan news detection task~\cite{kiesel-etal-2019-semeval} and the associated dataset, which inspired many publications.
Hyperpartisanship is defined as non-neutral news reporting~\cite{kiesel-etal-2019-semeval}, which can be described as a combination of linguistic and text-level biases on a document level.
The approach of \citet{jiang-etal-2019-team} performed best in the task.
It leveraged a convolutional neural network (\gls{CNN}) along with batch normalization and ELMo embeddings. 
In a follow-up study, \citet{jiang_comparing_2020} incorporated Latent Dirichlet Allocation (\gls{LDA}) distributions with different approaches to hyperpartisan news detection. 
They implemented multiple methods, such as a \gls{CNN}, a recurrent neural network (\gls{RNN}), a transformer encoder approach, and a hierarchical attention network (HAN) with and without \gls{LDA} topic modeling. 
Their results suggested that, in most cases, \gls{LDA} topic modeling improves the effectiveness of the methods, and hierarchical models outperform non-hierarchical models.
\citet{webson_undocumented_2020} presented another study based on the SemEval 2019 hyperpartisan news detection task.
They focused on decomposing pre-trained embeddings into separate denotation and connotation spaces to identify biased words descriptively. 
Although their primary goal was to improve the embeddings' reflection of the implied meaning of words, they showed how the discrepancy between the denotation space and the pre-trained embeddings reflects partisanship~\cite{webson_undocumented_2020}. 
\citet{cruz_document_2020} used different \gls{ML} approaches (e.g., \gls{RNN}, \gls{CNN}, bidirectional \gls{LSTM}/\gls{GRU}, and the attention-based approaches \gls{AttnBL}, \gls{HAN}) trained on the SemEval 2019 dataset.
They evaluated the effects of attention mechanisms and embeddings based on different granularities, tokens, and sentences on the effectiveness of the models. 
\citet{ruan_bias_2021} focused on introducing methods for generating additional data.
They presented two approaches for pseudo-labeling (overlap-checking and meta-learning) and introduced a system detecting media bias using sentence representations from averaged word embeddings generated from a pre-trained ELMo model and batch normalization.
The same authors also employed an ELMo-based classifier and a data augmentation method using pseudo-labeling~\cite{ruan_pseudo-labelling_2021}.

\textbf{Political Stance Detection.}
\citet{baly_we_2020} trained two models based on \gls{LSTM} and BERT for classifying news texts as left-wing, center, or right-wing. 
Their main contribution is the evaluation of techniques for eliminating the effects of outlet-specific language characteristics (here: political ideology expressed by linguistic bias) from the training process. 
They used adversarial adaptation and triplet loss pre-training for removing linguistic characteristics from the training data.
Further, they incorporated news outlets' Wikipedia articles and the bio of their Twitter followers in the training processes to reduce the effects of outlet-specific language characteristics.
While a transformer-based classification outperformed the \gls{LSTM} model, the techniques for improving training effectiveness improved both models' classification results.\footnote{Since transformers are not the paper's focus, we discuss it here.}
As part of their political stance detection approach, \citet{gangula_detecting_2019} proposed a headline attention network approach to bias detection in Telugu news articles. 
It leveraged a bidirectional \gls{LSTM} attention mechanism to identify key parts of the articles based on their headlines, which were then used to detect bias toward political stances.
They compared the results of their approach with \gls{NB}, \gls{SVM}, and \gls{CNN} approaches, all of which the headline attention network outperformed. 
To depolarize political news articles, \citet{fagni_fine-grained_2022} mapped Italian social media users into a 2D space. Their solution initially leveraged a \gls{NN} for learning latent user representations. Then, they forwarded these representations to a \gls{UMAP}~\cite{McInnes2018} model to project and position users in a latent political ideology space, allowing them to leverage properties of the ideology space to infer the political leaning of every user, via clustering.

\textbf{Gender/Group Bias.}
\citet{field_unsupervised_2020} presented an unsupervised approach for identifying gender bias in Facebook comments. 
They used a bidirectional \gls{LSTM} to predict the gender of the addressee of Facebook comments and, in doing so, identify gender biases in these comments.
\citet{mathew_hatexplain_2020} introduced HateXplain, a dataset on hate speech and gender bias that includes expert labels on the target community towards which the hate speech is aimed.
They further included labels of words annotators identified as bias-inducing.
They evaluated the effects of including the rationale labels in the training process of a \gls{BiRNN} and a BERT model on the models' bias detection capabilities.  
Including the rationale labels increased the bias classification performance for both models.

\subsubsection{Non-Neural Network Machine Learning Techniques}\label{sec:oML}
Besides state-of-the-art approaches using \gls{tbML} or deep learning techniques for bias detection, other (\gls{nNN}) \gls{ML} approaches are still widely used for bias detection. 
Many employ \gls{LDA}, \gls{SVM}, or regression models, but a wide range of models is usually used and compared. 
These models are particularly common in papers presenting new datasets, as they can be seen as a solid and widely known baseline for the quality of labels within a dataset.  

Based on the MBIC dataset, \citet{Spinde_2021a, Spinde_2021f} presented a traditional feature-based bias classifier. 
They evaluated various models (e.g., \gls{LDA}, logistic regression (\gls{LR}), \gls{XGBoost}, and others), trained with features such as a bias lexicon, sentiment values, and linguistic word characteristics (such as boosters or attitude markers~\cite{Spinde_2021a}).
\citet{alzhrani_ideology_2020} contributed a dataset of personalized news. 
Furthermore, she used a range of classifiers (Ridge classifier, nearest centroid, \gls{SVM} with SDG, \gls{NB}) for political affiliation detection.
\citet{rao2021gender} investigated coverage and gender bias in their dataset of Canadian news articles. 
They employed \gls{LDA} topic modeling to detect biased topic distributions for articles that contain predominantly male or female sources. 
\citet{kameswari_enhancing_2020} presented a dataset of 200 unbiased and 850 biased articles written in Telugu. 
They used \gls{NB} (Bernoulli and multinomial), \gls{LR}, \gls{SVM}, \gls{RF}, and \gls{MLP} classifiers to evaluate the effectiveness of adding presuppositions as model input. 
\citet{shahid_detecting_2020} researched framing effects in news articles using their proposed dataset. 
They trained an \gls{SVM} classifier to detect and classify moral framing and compared it to a baseline lexicon-based natural language processing approach, investigating moral framing aspects such as authority, betrayal, care, cheating, etc. 
\citet{ganguly_empirical_2020} explored various biases that can occur while constructing a media bias dataset. 
Part of their work examined the correlation between the political stance of news articles and the political stances of their media outlets.
To evaluate this correlation, they compared multinominal \gls{NB}, \gls{SVM}, \gls{LR}, and \gls{RF} models using ground-truth labels.
Several other publications described the application of \gls{nNN} \gls{ML} approaches in addition to other \gls{ML} techniques for data evaluation \citep{Lazaridou_2020, mokhberian_moral_2020, leavy_uncovering_2020, zhou_hidden_2021, orbach_out_2020}. 
We have already mentioned these in \Cref{sec:tNLP} and \Cref{sec:ntbML}.

\citet{baly_multi-task_2019} presented a multi-task ordinal regression framework for simultaneously classifying political stance and trustworthiness at different Likert scales. 
This approach is based on the assumption that the two phenomena are intertwined. 
They employed a copula ordinal regression along with a range of features derived from their previous work, including complexity and morality labels, linguistic features, and sentiment scores.
\citet{anthonio_team_2019} presented an additional\footnote{We mention multiple models for the task within \Cref{sec:ntbML}.} model for the SemEval 2019 hyperpartisan news detection task~\cite{kiesel-etal-2019-semeval}. 
They used a linear \gls{SVM} with \gls{VADER} sentiment scores as a feature, relying exclusively on the intensity of negative sentiment in texts to derive political stances expressed in texts.
With a $F_{1}$ score of 0.694, their approach failed to match the other competitors in the task.
In addition to a FastText classifier, the approach presented by \citet{Lazaridou_2020} included a manual selection of training data containing examples of media bias. 
Aside from contributing to a new media bias dataset and evaluating the effect of expert and non-expert annotators, they presented a curriculum learning approach for media bias detection.
They concluded that high-quality expert-labeled data improves the performance of the model. 

\subsection{Graph-Based}\label{sec:Graph-based}
The research described in this section leverages graph data structures to analyze online social networks through their users and text interactions, which requires a distinctive set of methods for bias analysis.
Although most publications used \gls{ML}, we treat them separately due to the unique characteristics of the analyzed data representations.
Graph-based approaches are primarily used to investigate framing bias, echo chambers, and political stances.
Therefore, we structure our overview of corresponding publications by the type of bias they investigate.

\textbf{Framing Bias.}
The SLAP4SLIP framework~\cite{hofmann_modeling_2021} detects how concepts are discussed in different parts of a social network with predefined linguistic features, graph \gls{NN}, and structured sparsity.
The authors exploit the network structure of discussion forums on Reddit without explicitly labeled data and minimally supervised features representing ideologically driven agenda setting and framing.
Training graph auto-encoders, \citet{hofmann_modeling_2021} modeled agenda setting, and framing for identifying ideological polarization within network structures of online discussion forums. They modeled polarization along the dimensions of salience and framing. Further, they proposed MultiCTX (Multi-level ConTeXt), a model consisting of contrastive learning and sentence graph attention networks to encode different levels of context, i.e., neighborhood context of adjacent sentences, article context, and event context.

\citet{guo_modeling_2022} built on the SLAP4SLIP framework~\cite{hofmann_modeling_2021} to detect informational bias and ideological radicalization by combining contrastive learning and sentential graph networks.
Similarly, \citet{tran_how_2020} proposed a framework for identifying bias in news sources.
The authors used BERT Base for aspect-based sentiment analysis and assigned a bias score to each source with a graph-based algorithm.

\textbf{Echo Chambers.}
\citet{villa_echo_2021} applied community detection strategies and modeled a COVID-19-related conversation graph to detect echo chambers.
Their method considered the relationship between individuals and the semantic aspects of their shared content on Twitter.
By partitioning four different representations of a graph (i.e., topology-based, sentiment-based, topic-based, and hybrid) with the METIS algorithm\footnote{As proposed by \citet{Karypis_1995}.}, followed up by qualitative methods, they assessed both the relationships connecting individuals and semantic aspects related to the content they share over Twitter. 
They also analyzed the controversy and homogeneity among the different polarized groups obtained.

\textbf{Political Stance Detection.}
Stance detection\footnote{We defined stance detection as political bias detection via the identification of linguistic biases, compare \Cref{sec:TransformerBasedModels}.} is a typical application of graph-based classification techniques. 
\citet{zhou_uncovering_2021} combined network structure learning analysis and \gls{NN} to predict the political stance of news media outlets.
With their semi-supervised network embedding approach, the authors built a training corpus on network information, including macro- and micro-network views.
They primarily employed network embedding learning and graph-based label propagation to overcome label sparsity.
By integrating graph embeddings as a feature, \citet{stefanov_predicting_2020} detected the stance and political stance of Twitter users and online media by leveraging their retweet behavior.
They used a user-to-hashtag graph and a user-to-mention graph and then ran node2vec.
They achieved the best result for combining BERT with valence scores\footnote{A valence score~\cite{stefanov_predicting_2020} close to zero reflects that an influencer is cited evenly among different groups in a network. 
Conversely, a score close to $-1$ or 1 indicates that one group disproportionately cites an influencer compared to another group. 
In their paper, \citet{stefanov_predicting_2020} indicated that valence scores are essential in identifying media bias in social networks.}.
\citet{guimaraes_characterizing_2021} analyzed news stories and political opinions shared on Brazilian Facebook.
They proposed a graph-based semi-supervised learning approach to classify Facebook pages as politically left or right.
Utilizing audience interaction information by inferring self-reported political leaning from Facebook pages, \citet{guimaraes_characterizing_2021} built an interest graph to determine the stance of media outlets and public figures.
The authors achieved the best results for label propagation with a spectral graph transducer.
\citet{li-goldwasser-2019-encoding} captured social context with a neural architecture for representing relational information with graph-based representations and a graph convolutional network. 
They showed that using social information, such as Twitter users who have shared the article, can significantly improve performance with distant and direct supervision.

\subsection{Bias in Language Models}
Detecting bias inherent to language models is an important research area due to the models' popularity for many NLP tasks. 
Researchers have investigated bias in texts and other media generated by language models as well as in classification performed with language models.
We did not include publications that address these forms of bias.\footnote{We focus exclusively on detection methods; the field of bias in language models is extensive enough for a dedicated literature review.} 
However, we would like to give some examples to raise awareness of biased language models.
\citet{nadeem_stereoset_2021} analyzed stereotypical bias with the crowdsourced dataset StereoSet in BERT, GPT-2, ROBERTA, and XLNET, concluding that all models exhibit strong stereotypical bias.
\citet{vig_investigating_2020} used causal mediation analysis to analyze gender bias in language models.
Their results showed that gender bias effects exist in specific components of language models.
\citet{bhardwaj_investigating_2021} also analyzed gender bias within BERT-layers and concluded that the layers are generally biased.
In \citet{Liu2021MitigatingPB}, the authors detected bias in texts generated by GPT-2 and discussed means of mitigating gender bias in language models by using a reinforcement learning framework.

\subsection{Datasets}\label{sec:datasets}
During our review, we collected both methods and datasets from the publications we selected for inclusion. In total, we found 123 datasets.  
We categorize the datasets according to the concepts proposed in our Media Bias Taxonomy, similar to the discussion of methodologies as shown in \Cref{table:datasets}.
We added the category General Linguistic Bias as several datasets do not define the subcategory of bias they contain.
We did not evaluate the quality of the datasets as they address distinct tasks and objectives but leave this assessment for future work (cf. \Cref{sec:discussion}). 

Only two of the 123 datasets include information on the background of annotators. Moreover, dataset sizes are generally small; only 21 of the 123 datasets contain more than 30,000 annotations. 
We believe that the use of multiple datasets is promising for future work as we discuss in \Cref{sec:discussion}. 
As part of this review, we present the datasets,  their statistics, and tasks merely as a starting point for future work, without further assessment. 
We give a detailed overview of publications, sizes, availability, tasks, type of label, link, and publication summary for each dataset in our \hyperref[taxonomyurl]{repository}.

\begin{table}[H]
\begin{tabular}{|l|l|l|}
\hline 
Media Bias Category & Media Bias Type & Amount \\ \hline Linguistic Bias              &                             &   45                                                                                     \\ \hline
                             & General Linguistic Bias & 26                                                                                        \\ \hline
                             & Framing Bias                & 15                                                                                        \\ \hline
                             & Epistemological Bias        & 3                                                                                        \\ \hline
                             & Bias by Semantic Properties & 1                                                                                                                                                                             \\ \hline
Text-level Context Bias      &                             & 5                                                                                         \\ \hline
                             & Statement Bias              & 2                                                                                        \\ \hline
                             & Phrasing Bias               & 3                                                                                        \\ \hline
Reporting-level Context Bias &                             & 6 
\\ \hline
                             & General Reporting-level Context Bias              & 2                                                                                        \\ \hline
                             & Selection Bias              & 1                                                                                        \\ \hline
                             & Coverage Bias               & 2                                                                                        \\ \hline
                             & Proximity Bias              & 1                                                                                        \\ \hline
Cognitive Bias.              &                             & 28                                                               \\ \hline
                             & Partisan Bias               & 28                                                        \\ \hline
Related Concepts             &                             &                                                              \\ \hline
                             & Hate Speech                 & 14                                                                                       \\ \hline
                             & Group Bias                  & 20                                  \\ \hline
                             & Sentiment Analysis          & 10                                                                                       \\ \hline
\end{tabular}
\caption{Overview of datasets found during our literature review} 
\label{table:datasets}
\end{table}

\section{Human-centered research on media bias} \label{sec:psychology}

Human-centered research on media bias aims to understand why people perceive media as biased, explore the societal and digital consequences, and develop strategies to overcome biased perception and detect media bias. Debates on all these factors are ongoing and experimental effects tend to be minor. Hereafter, we highlight some of these debates.

\subsection{Reasons for biased media perception}
One explanation for the emergence of cognitive biases in media perception is that information is processed in light of prior expectations, which may be distorted \citep{10.1037/1089-2680.2.2.175}. The veracity of claims is often judged based on familiarity, potentially resulting in illusory truths \citep{https://doi.org/10.1111/j.1460-2466.2009.01452.x, https://doi.org/10.1002/ejsp.264, doi:10.1177/0093650209333030, https://doi.org/10.1111/j.1751-9004.2007.00060.x}. Cognitive dissonance theory posits that people experience discomfort when confronted with information inconsistent with their convictions, motivating them to discount it \citep{festinger1957theory}.

Extending this notion to groups, \citet{tajfel1979integrative} suggested in their social identity and categorization theory that basic self-esteem is derived from personal affiliation with positively-connotated groups. This results in in-group favoritism, out-group derogation \citep{https://doi.org/10.1111/j.2044-8309.1989.tb00866.x, doi:10.1177/0146167206293190}, and behavior and information processing in line with group identity. People easily regard reports that negatively affect groups they strongly identify with as a personal threat to their self-esteem and devalue these reports \citep{doi:10.1080/19331681.2014.997416, 10.1111/jcom.12031}. Furthermore, \citet{Turner1991} posited that when people self-categorize with a specific group, they evaluate the validity of arguments by congruence to in-group norms and in-group consensus. This pattern aligns with empirical findings showing that news acceptance depends on group identification and congruent group membership cues of the news source \citep{doi:10.1177/0093650218794854, reid2012self-categorization}.

Generally, prior works expect selective exposure to media to be consistent with previous viewpoints \citep{klapper1960effects}, further strengthening prior convictions. Such behaviour can be referred to as confirmation bias \citep{10.1037/1089-2680.2.2.175} through repeated exposure \citep{10.1177/1088868309352251}. In the age of social media and the abundance of information available, these cognitive biases may further allow for confrontation only with attitude-consistent information and like-minded individuals in echo chambers \citep{Sunstein2009, 10.1002/asi.24121, ojs.aaai.org/index.php/ICWSM/article/view/14429}. Moreover, algorithms trained on these biases may further limit the available media spectrum in filter bubbles \citep{Pariser2011}.

Consequently, limited exposure to alternative viewpoints may also impact the perception of social norms and the prevalence of opinions. The overestimation of the frequency of one's own position, known as the false consensus effect \citep{10.1016/0022-1031(77)90049-X}, has been widely documented even before the introduction of social media and may be partially due to identity motivations explained earlier \citep{10.1037/0033-2909.102.1.72}. However, when echo chambers are used to gauge the frequency of opinions and social norms, even larger shifts between groups are expected \citep{10.1371/journal.pone.0147617}. This feeds into a vicious circle of polarizing group norms, discounting information inconsistent with these shifted norms, and feeling encouraged to voice even more extreme positions (e.g., \citep{Turner1991, 10.1177/0093650217745429, doi:10.1177/0093650218794854, http://doi:10.7910/DVN/TJKIWN}). These mechanisms lead to expectations that media perception is polarized based on social categories and prior beliefs and that the introduction of social media has exacerbated this phenomenon.

\subsection{Consequences of biased media perception}

Partisan individuals tend to select media that aligns with their prior beliefs and political attitudes, a phenomenon known as the Friendly Media Phenomenon (\gls{FMP}) \citep{doi:10.1080/10584609.2010.544280, doi:10.1177/0093650209333030, doi:10.1177/0093650217713066}. This tendency may be partially due to interpersonal communication among like-minded individuals \citep{doi:10.1177/0093650220915041}. People also tend to assess the veracity of information based on its fit with their political convictions, exhibiting partisan bias \citep{10.1016/j.tics.2021.05.001}.

Biased media perception can lead to the Hostile Media Phenomenon (\gls{HMP}), where people perceive media coverage as biased against their side, regardless of the actual political position of the article \citep{doi:10.1080/15205436.2015.1051234, Vallone_1985, https://doi.org/10.1080/08824096.2011.565280}. This effect increases with the extremity of party affiliation and is primarily due to the derogation of dissenting media \citep{Spinde2022a, doi:10.1177/0093650217713066, gearhart2020hostile}, making it a cognitive bias rather than a characteristic of the media landscape. Discussions and feedback from like-minded individuals can further amplify the \gls{HMP}, leading to the perception of general media bias even when primarily exposed to self-selected, like-minded media \citep{CastroHopmannNir+2020, doi:10.1177/0093650219836059}.

Methodologically, the \gls{HMP}, \gls{FMP}, and partisan bias complicate the assessment of media bias, as raters' perceptions of bias may reflect more on individual affiliations and idiosyncrasies than the objective properties of the rated article \citep{Spinde2021e}. Subjective bias ratings are relative to their social context; their quality as a scientific measure of media bias depends on the representativeness of raters. Therefore, such ratings should be supplemented by objective bipartisan bias criteria (e.g., language biases).

Socially, the \gls{HMP} can lead to the mobilization of more extreme positions, distrust in the social system, and, in cases of low efficacy beliefs, political withdrawal \citep{doi:10.1080/15205436.2015.1051234}. Both the \gls{HMP} and \gls{FMP} can contribute to increased political segmentation and polarization, which can negatively impact political communication and interaction, essential for a peaceful and democratic society \citep{gearhart2020hostile}. Exposure to certain media can also have social consequences, such as altered political participation \citep{doi:10.1177/0093650219836059}. For example, \citet{doi:10.1177/0093650215593145} found that exposure to congruent media is tied to biased perceptions of the opinion climate, influencing how participants communicate their political beliefs and engage in politically meaningful acts, while incongruent exposure has little effect.

The role of the social media environment in this process is somewhat disputed: While selective exposure in social media is widely documented \citep{10.1002/asi.24121, 10.1073/pnas.1517441113}, some authors argue that social media is not the main contributor to the variety of media diets globally. For example, \citet{10.14763/2016.1.401} deem its general impact negligible and suggest it may expose users to more diverse information compared to traditional media. According to \citet{Dubois_2018}, people may even cope with this high-choice media environment by developing strategies like verifying news in different outlets, and---even though social networks are polarized---only a subset of the population regards itself as susceptible to echo chambers. After all, the phenomena and underlying cognitive processes were known before the advent of social media. The effects observed in social media may just be more visible to researchers than they were before \citep{10.14763/2016.1.401}. In addition, exposure to biased media may not be sufficient to significantly affect attitudes \citep{10.1038/s41591-022-01713-6}. As such, it is challenging to determine the overall effect of social media on biased media perception and social consequences today, though some feedback loops can be expected \citep{http://doi:10.7910/DVN/TJKIWN}. This problem is even more pressing for algorithmic filtering than for personal selections, as the algorithms involved are not transparently disclosed, their application is in flux, and they are not accessible to the user \citep{10.14763/2016.1.401}. This fact illustrates that parts of the conclusion on the impact of social media on media bias phenomena are also driven by the selection of media and the assessment method of the effects.

\subsection{Recipient-oriented approaches to reduce media bias}


Given that selective media exposure partially explains cognitive media bias phenomena, one intervention approach is to encourage and facilitate a diverse media diet to reduce media bias \cite{Dubois_2018}. This can be achieved by plug-ins that actively diversify the media displayed in a search by identifying the topic and sampling other articles or information related to it \citep{10.1145/3311957.3359513}, or by providing media based on another individual's platform history \citep{10.1145/3491102.3502028}. In a similar approach, \citet{ojs.aaai.org/index.php/ICWSM/article/view/14429} used a browser widget to provide feedback on the balance of a user's media diet, successfully encouraging these users to explore more media from centrist and opposing viewpoints.

Other experiments and observations of counter-attitudinal exposure illustrate that the mere presentation and reception of opposing viewpoints do not always decrease the \gls{HMP} and may even exacerbate the problem. For instance, \citet{10.1111/jcc4.12199} found that people who were incidentally exposed to counter-attitudinal information are more likely to subsequently select information that aligns with their attitudes. Other studies found that exposure to incongruent comments increases the perception of bias and decreases the perception of the credibility of a later, neutral news report \citep{gearhart2020hostile}, and that exposure to opposing tweets may backfire and intensify political polarization, particularly for Republicans \citep{Bail9216}. These findings are consistent with the notion of motivated reasoning, as the potential threat of backfiring from inconsistent exposure---though rather dependent on the specific materials to which readers are exposed \citep{10.1016/j.jarmac.2020.06.006}---may be explained by the threat of the presented material to the reader's identity. As a result, diverse exposure with well-crafted materials may help but is not a comprehensive solution for the \gls{HMP}, \gls{FMP}, and biased media perception.


As an alternative, some studies have attempted to alter the user's mindset during news processing and shift the attentional focus to aspects of a user's self-identity that are not challenged by the news report. For example, inducing self-affirming thoughts aimed at mitigating the potentially self-threatening aspect of belief-inconsistent arguments has been shown to successfully evoke more unbiased processing of such information \citep{10.1016/j.jesp.2003.07.001}. Similarly, focusing readers' attention on a value that may be threatened by information increases their perception of media bias in that article \citep{doi:10.1080/08824096.2018.1555659}. Likewise, people seem more open to sharing and are better at judging news headlines based on their veracity when nudged to think about their own accuracy instead of their identity motives \citep{10.1016/j.tics.2021.02.007}. Opening the mindset may thus be an effective, albeit situational, approach when tackling phenomena such as the \gls{HMP} and media bias detection during exposure to attitude-inconsistent materials.


As an additional step, forewarning messages that draw attention to biased media and potential influencing attempts can help ``inoculate'' against this media by provoking reactance towards manipulations \citep{Spinde2022a, 10.1017/bpp.2020.60, 10.1098/rsos.211719}. Exposing individuals to examples of media bias through such messages may teach them to detect and cope with it. In this vein, various forms of training have been tested and generally increase a reader's ability to identify biased media and distinguish it from congruency with one's political stance \citep{10.1038/s41591-022-01713-6, Spinde2022a, Spinde_2020}. This detailed training is necessary, as mere awareness of media bias as part of general news media literacy may not be sufficient for a balanced media diet \citep{doi:10.1177/1464884918805262}.

Overall, all approaches have yielded relatively small effects on improving media bias detection, and more research on effective interventions is necessary. Regarding partisan bias, there is some indication that interventions are not equally effective in reducing the bias for liberals and conservatives---potentially inadvertently biasing the overall discourse on media towards the less open-minded faction \citep{10.25384/SAGE.12594110.v2, Spinde2022a}. Thus, further testing of the effectiveness of approaches in reducing partisan media perception and the \gls{HMP} is warranted.

\section{Discussion} \label{sec:discussion}

To address RQ1, we have established a Media Bias Taxonomy that allows to precisely categorize the various sub-concepts related to media bias \cite{Spinde_2021f, Spinde2021e}. We emphasize the complexity of media bias and note that researchers often fail to clearly define the type of media bias they investigate, which leads to confusion when comparing different studies. Furthermore, existing literature reviews on the topic do not address the various media bias concepts \cite{Hamborg2019inter}, making it difficult to understand problems and solutions across different approaches.

Our Media Bias Taxonomy is a crucial first step in establishing a common ground for more clearly defined media bias research. We divide media bias into five major categories: linguistic bias, cognitive bias, text-level context bias, reporting-level bias, and related concepts. We provide subgroups for each of these categories. Throughout the creation of our taxonomy, we engaged in frequent discussions and revised our definitions and structure multiple times, revealing the numerous options available for defining media bias.

While our taxonomy provides a practical foundation and effective starting point for research in the domain, future research should critically re-examine the discussed concepts. We believe that the main common ground among the various types of media bias we identified is smaller than that of existing universal definitions (see \Cref{sec:bias}) and primarily refers to one-sided media content

To answer RQ2 and RQ3, we provided an extensive overview of recently published literature on computer science methods and datasets for media bias detection. We manually inspected over 1,528 computer science research papers on the topic published between 2019 to May 2022 after automatically filtering over 100,000 keyword-related publications. Our review reveals valuable insights into best practices and trends in the research field.

In recent years, transformers have quickly become the most frequently used and most reliable method for media bias detection and debiasing \cite{Spinde_2021f}. Platforms like Hugging Face facilitate the implementation of the models and their adaption to various tasks \cite{dinkov_predicting_2019}. However, as we show in \Cref{csmethods}, the new models have not yet made their way into all subtypes of bias, leaving room for future experiments. Additionally, available media bias classifiers are largely based on small in-domain datasets. Recent advancements in natural language processing, especially transformer-based models, demonstrate how accurate results can be achieved by unsupervised or supervised training on massive text corpora \cite{DBLP:journals/corr/abs-2111-10952} and by model pre-training using inter and cross-domain datasets \cite{DBLP:journals/corr/abs-2111-10952}.

Although graph-based methods are not as popular as transformers, their application to media bias detection is increasing but mostly limited to analyzing social network content, activities, and structures, and identifying structural political stances within these entities \cite{stefanov_predicting_2020, guimaraes_characterizing_2021, zhou_uncovering_2021, li-goldwasser-2019-encoding}. Transformer-based approaches cannot accomplish such an analysis due to the network properties of the explored data.

Established methods still play a role in media bias detection. Traditional natural language processing approaches, as well as non-transformer-based (deep \gls{NN}) machine learning models, are simpler and more explainable compared to language-model-based approaches, making them advantageous in applications where transparency of classification decisions is critical (e.g., \cite{Spinde_2021a}). Since traditional approaches have been used in many media bias identification tasks, they often serve as a baseline to compare new (transformer-based) approaches. Given their higher explainability and long-term testing, we don't expect language models to completely replace other approaches soon.

Apart from these major trends, including information on spreading behavior, social information \cite{chitra_analyzing_2020, stefanov_predicting_2020, li-goldwasser-2019-encoding}, metadata \cite{dinkov_predicting_2019}, and examining the vector spaces of word embeddings \cite{dinkov_predicting_2019} also show promise in improving classifier performance to detect media bias.

We addressed RQ4 by reviewing social science research on media bias. One significant takeaway is that media bias datasets largely ignore insights from social science research on the topic, leading to low annotator agreement and less accurate annotations \cite{Spinde_2021f}. The perception of bias depends on factors beyond content, such as the reader's background and understanding of the text. Moreover, limited exposure to alternative viewpoints can impact how social norms and opinions are perceived. These insights have never been fully integrated into automated detection methods or datasets. Integrating bias perception research in language models is a promising way to improve annotation-based detection systems \cite{bhardwaj_investigating_2021}, which can potentially be achieved by further developing standardized questions within the domain \cite{Spinde_2021f}.

We see a need to develop further methods to increase news consumers' bias awareness and believe that computer science methods, as described in this review, can be a powerful tool to build such awareness-increasing tools. While some tools already exist, none have been applied on a larger scale in a real-world scenario, which is a promising direction for future research.

Our literature review also exhibits limitations. First, we excluded work from areas other than media bias due to the high number of publications involved, potentially leaving out valuable contributions. Investigating promising concepts from other areas will be necessary for future work. Second, for all computer science methods, we only included literature from 2019 to 2022, excluding valuable earlier research. Analyzing a longer period could yield an even more complete picture of the research domain. Lastly, although we distinguish several categories within our Media Bias Taxonomy, the concepts related to media bias still overlap and appear concurrently. We believe that future work should further discuss and adapt the taxonomy. Although the taxonomy we present is merely a starting point to connect works in the area, we believe it can benefit future approaches by raising awareness of concepts, methods, and datasets in the research domain. During the writing of this literature review, the taxonomy's outline frequently changed in permanent discussions among the authors.

\section{Conclusion}\label{sec:conq}

In 2018, \citet{Hamborg2019inter} concluded that (1) powerful computer science methods (such as word embeddings and deep learning) had not yet made their way into the automated detection of media bias and that (2) the interdisciplinarity of media bias research should be improved in the future. The authors suggested (3) that approaches in computer science did not account for bias having many different forms and usually only focused on narrow bias definitions \cite{Hamborg2019inter}.

Our literature review reveals that two of these propositions (1 and 3) have been addressed to some extent, but there is still considerable room for improvement. Transformer and graph-based methods have led to significant increases in the performance of automated methods for detecting media bias, and numerous types of bias have received research attention. However, these concepts are primarily used and analyzed individually, with knowledge overlaps between them remaining unexplored \cite{spinde_interdisciplinary_2021}. Recent modeling techniques, such as multi-task learning, enable the use of related datasets to improve classification performance \cite{DBLP:journals/corr/abs-2111-10952}.

Regarding (2), datasets and systems still exhibit limited conceptual work, with the cognitive dimension of media bias rarely mentioned in computer science research. Our literature review aims to provide a foundation for increased awareness of bias in media bias datasets (through standardized annotator background assessments), enhanced interdisciplinarity in the research domain (which we believe is particularly relevant since reasonable classifications cannot exist without clear conceptualizations), and future computer science methods.

We are confident that this review will facilitate entry into media bias research and help experienced researchers identify related works. We hope that our findings will contribute to the development of more effective and efficient media bias detection methods and systems to increase media bias awareness. Finally, we plan to repeat our workflow in three years to reassess the state of the research domain.

\begin{acks}
We thank Elisabeth Richter, Felix Blochwitz, Jerome Wassmuth, Sudharsana Kannan, and Jelena Mitrović for supporting this project through fruitful discussions. We are grateful for the financial support of this project provided by the Hanns-Seidel Foundation, the DAAD (German Academic Exchange Service), the Lower Saxony Ministry of Science and Culture, and the VW Foundation.
\end{acks}

\bibliographystyle{ACM-Reference-Format}
\bibliography{joined}

\printglossaries

\newpage
\appendix
\section{Appendices} \label{sec:appendix}

\subsection{Publications per Bias Category}
\label{app:2}
\begin{figure}[H]
    \centering
    \includegraphics[width=\linewidth]{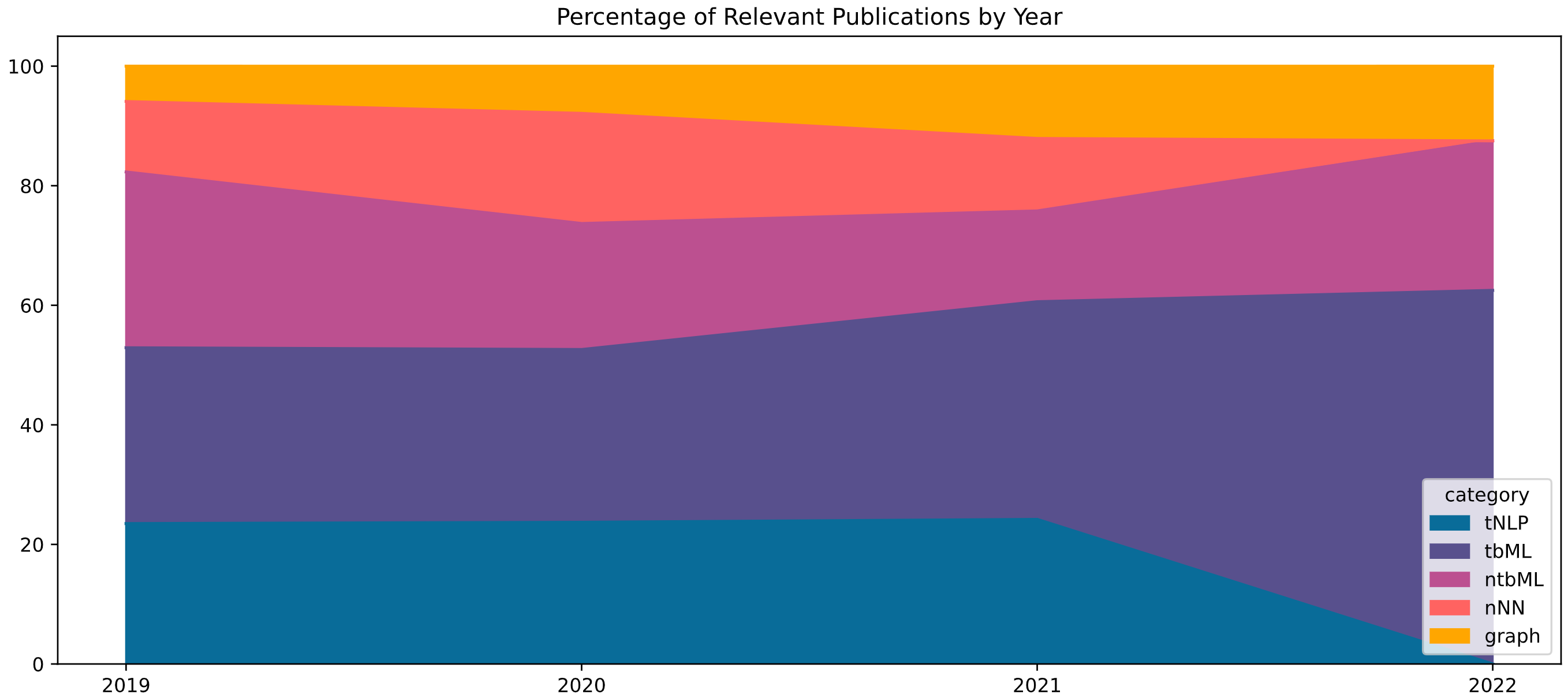}
    \caption{Overview of relevant publications per category and year. Publications can appear in multiple categories.}
\end{figure}

\subsection{Definitions and Examples of Media Bias}
  \label{app:1}

%
%
%
%
%
\begin{longtable}{P{0.13\textwidth}P{0.13\textwidth}p{0.28\textwidth}P{0.05\textwidth}P{0.28\textwidth}}
\multicolumn{5}{l}%
{{\bfseries Linguistic Bias Overview}} \\
\hline
Subcategory &
  Features &
  Definition &
  Source &
  Example \\ \hline
  \endfirsthead
  
  \hline
Subcategory &
  Features &
  Definition &
  Source &
  Example \\ \hline
  \endhead
Framing Bias &
   &
  Subjective words or phrases linked with a certain point of view. &
  \citep{Recasens_2013} &
  Usually, smaller cottage-style houses have been demolished to make way for these \textbf{\textit{McMansions}}. \\
 &
  Subjectivity Clues &
  Subjectivity clues are words that, in most (\textit{strongsubj}) or certain (\textit{weaksubj}) contexts, are considered to add subjective meaning to a statement. &
  \citep{riloff-wiebe-2003-learning, wilson-et-al2005recognizing} &
  Subjective adjectives (e.g., preposterous, unseemly) and metaphorical or idiomatic phrases (e.g., dealt a blow, swept off one's feet). \\
 &
  One-sided Terms  &
  One-sided terms only reflect one side of a controversial subject where the same event can be seen from two or more opposing perspectives. &
  \citep[p. 4]{Recasens_2013} &
  ``\textbf{\textit{Pro-life}}'' vs. ``anti-abortion''. \\
Epistemological Bias  &
   &
  Linguistic features that subtly focus on the believably of a position. &
  \citep{Recasens_2013} &
  Kuypers \textbf{\textit{claimed}} that the mainstream press in America tends to favor liberal viewpoints. \\
 &
  Factive Verbs &
  Factive verbs presuppose the truth of a complement clause. &
  \citep{hooper1975on}, \citep[p. 5]{karttunen1971implicative} &
  ``John \textbf{\textit{realized}} that he had no money`` implies ``John had no money''. \\
 &
  Entailments &
  Entailments are directional relations where one statement implies the truth of another. &
  \citep[p. 1]{berant-etal-2012-efficient}, \citep{karttunen1971implicative} &
  ``Cyprus \textbf{\textit{was invaded by}} the Ottoman Empire in 1571'' implies the hypothesis ``The \textbf{\textit{Ottomans attacked}} Cyprus''. \\
 &
  Assertive Verbs &
  Assertives imply that the speaker or subject of a sentence has an affirmative opinion on the truth value of a complements proposition. &
  \citep{hooper1975on} &
  E.g., regret, resent, forget, amuse, suffice, bother. \\
 &
  Hedge Words &
  Hedge words are used to lessen the probability and the writer's guarantee of the truthfulness of a statement. &
  \citep{hyland2005metadiscourse} &
  E.g., perhaps, might, may. \\
Bias by Semantic Properties &
   &
  A set of semantic properties which are grammatically relevant and serve as an interface between syntax and semantics. &
  \citep[p. 3]{Greene_2009} &
  ``The gunmen shot the opposition leader'' vs. ``The shooting killed the opposition leader''. \\
 &
  Causative-Inchoative Alternations &
  Causative-inchoative alternations are verbs that can either emphasize a change of state or the bringing about of the change of state. &
  \citep{Greene_2009, Levin1993}, \citep[p. 1]{Pin2001AFL}&
  ``Rebecca broke the pencil'' vs. ``The pencil broke''. \\
Connotation Bias \label{ConnotationBias} &
   &
  Bias through the use of different words for the same subject with identical denotations but differing (often biased) connotations. &
  \citep{webson_undocumented_2020, chakrabarty2021entrust} &
  E.g. ``immigrants'' vs. ``\textbf{\textit{aliens}}'', ``estate tax'' vs. ``\textbf{\textit{death tax}}''. \\
 &
  Verb Connotations &
  Verbs with identical denotation but differing connotations. &
  \citep{Rashkin_2016} &
  The story begins in Illinois in 1987 when a 17- year-old girl \textbf{\textit{suffered}} a botched abortion. \\
 &
  Noun and Adjective Connotations &
  Nouns and adjectives with identical denotation but differing connotations. &
  \citep{allaway2021unified} &
  relentless (-) vs. persistent (+) gentleman (+) vs. man (=) protection (-) vs. security (=)

  \\
Linguistic Intergroup Bias &
   &
  Variations in language lead to the creation or maintenance of the reputations of certain social groups. &
  \citep[p. 2]{Dragojevic_2016} &
  ``Mary hit Mike'' vs. ``Mary hurt Mike'' vs. ``Mary hates Mike'' vs. ``Mary is aggressive''. \\ \hline
\caption{Linguistic Bias Overview} 
\label{table:categories1}
\end{longtable}

\begin{longtable}{P{0.16\textwidth}p{0.34\textwidth}P{0.05\textwidth}P{0.34\textwidth}}
\multicolumn{4}{l}%
{{\bfseries Context Bias (Text Level)}} \\
\hline
Subcategory &
  Definition &
  Source &
  Example \\ \hline
  \endfirsthead
  
  \hline
Subcategory &
  Definition &
  Source &
  Example \\ \hline
  \endhead
Spin Bias &
  Emerges through a newspaper's attempt to create a memorable story by means of simplification or exaggeration. In the case of simplification, newspapers discard some information, whereas exaggeration shapes the story in a certain direction. Both methods try to present the news in a way to attract potential readers. &
  \citep{Mullainathan_2002, alsem2008impact} &
  President Donald Trump \textbf{\textit{gloated}} over mass layoffs at multiple news outlets on Saturday. \\
Statement Bias &
  Refers to the process of how facts are reported. This can be by means of expressing own opinions, criticizing the counterpart, or advocating the own viewpoints in a way that news reporting is favorable or unfavorable towards a certain standpoint. &
  \citep{D'Alessio_2000} &
  A political protest in which people sat in the middle of a street blocking traffic can be described as ``peaceful'' and ``productive,'' or others may describe it as ``aggressive'' and ``disruptive''. \\
Omission Bias / Commission Bias &
  Omission or Commission Bias results from the journalists deciding what (substantial) information to exclude and what information to include in the news report. &
  \citep{Alonso_2017, Hamborg2019inter} &
  CNN previously reported on the FBI's hate crime statistics released last November, which showed the number of hate crimes reported to the bureau rose by about 17\% in 2017 compared to 2016. 2017 is the latest year for which those statistics are available. It was the third-straight year that hate crime incidents rose.\footnote{Example from https://www.allsides.com/media-bias/how-to-spot-types-of-media-bias} \\
Informational Bias &
  A statement containing Informational Bias influences the readers' opinion by adding information that is not necessarily relevant to the event or is rather speculative. &
  \citep{fan_plain_2019} &
   Looking at two articles that report on the same event, the Huffington Post and Fox News each frame entities of opposing stances negatively. HPO states an assumed future action of Donald Trump as a fact, and FOX implies Democrats are taking advantage of political turmoil.
 \\
Phrasing Bias &
  Phrasing Bias results from using words in a sentence or statement which are inflammatory or partial. &
  \citep[p. 2]{Hube_2018b} &
  ``Aborting the fetus'' vs. ``\textbf{\textit{Killing the baby}}''.
  \\ \hline
  \caption{Context Bias (Text Level)} 
\label{table:categories2}
\end{longtable}

\begin{longtable}{P{0.16\textwidth}p{0.34\textwidth}P{0.05\textwidth}P{0.34\textwidth}}
\multicolumn{4}{l}%
{{\bfseries Context Bias (Reporting Level)}} \\
\hline
Subcategory &
  Definition &
  Source &
  Example \\ \hline
  \endfirsthead
  
Coverage Bias &
  Coverage bias is concerned with the different amounts of coverage diverse sides of the same issue receive. &
  \citep{D'Alessio_2000} &
  A candidate of a political party receiving more coverage than the candidate of the other party. \\
Selection / Gatekeeping Bias &
  Selection bias occurs because individuals in the news media are biased in their selection of events to report on. &
  \citep{D'Alessio_2000} &
  During the Roosevelt administration, the entire press was hostile to it in some parts of the country, and it was impossible to get the other side of the story.
  \\ \hline
    \caption{Context Bias (Reporting Level)} 
\label{table:categories3}
\end{longtable}

\begin{longtable}{P{0.16\textwidth}p{0.34\textwidth}P{0.05\textwidth}P{0.34\textwidth}}
\multicolumn{4}{l}%
{{\bfseries Cognitive / Perception Bias}} \\
\hline
Subcategory &
  Definition &
  Source &
  Example \\ \hline
  \endfirsthead
  
  \hline
Subcategory &
  Definition &
  Source &
  Example \\ \hline
  \endhead
Political Ideology &
  Political Ideology determines the likelihood of a news consumer perceiving media as biased and influences what news sources are viewed as biased. &
  \citep{glynn2014how, lee2005liberal} &
  Viewers of Fox News held a distinct set of attitudes towards President George W. Bush (i.e., more favorable) as to his political opponents (i.e., less favorable). \\
Hostile Media Phenomenon &
  The tendency of partisans to view media coverage as negative towards their own positions. Partisans also fear that nonpartisans could be swayed into a hostile direction by the media coverage. &
  \citep{Vallone_1985, reid2012self-categorization, guntherschitt2004mapping, guntherliebhardt206broad, schmitt-et-al2004why, dalton-et-al1998american, Spinde_2021a} &
  Both Pro-Arab and pro-Israeli subjects who saw US news programs on the  Israeli move into West Beirut perceived it as biased in favor of the other side. \\
Interpersonal Factors &
  Interpersonal Factors (e.g., ideological similarity) are related to the perception of media bias. &
  \citep{eveland2003impact, CastroHopmannNir+2020} &
  Conversations with ideologically like-minded others increase the readers' likelihood of perceiving media as biased. \\
Article Comments &
  A news consumer being exposed to other users' comments may influence the readers' perception of this article being biased. Research shows that, specifically, those individuals who are confronted with incongruent opinions are likely to perceive stronger levels of media bias. &
  \citep{houston2003influence, lee2012thats, gearhart2020hostile} &
  Readers who saw a neutral Facebook post of an article on abortion with comments in line with their opinion perceived the story, writer, and outlet as less biased against their opinion. \\
Exposure to Opposing Views &
  Constant exposure to messages with opposing views from politicians and opinion leaders leads media consumers to perceive more bias in news media. &
  \citep{Bail9216} &
  Republicans who followed Twitter bots that retweeted liberal content exhibited more conservative views after one month of exposure. \\
Topic Identification / Involvedness &
  High involvement or identification with a topic or an issue is a reason consumers perceive media to be biased. &
  \citep{zhang2021effects, glynn2014how} &
  Persons who strongly identified with the republican party and were pro-gun ownership perceived media bias in mass shooting news against gun owners to a higher degree. \\
Outlet Reputation &
  The reputation of media outlets' ideological orientation can lead consumers to perceive bias in balanced news. &
  \citep{baum2008intheeye} &
  The perception of bias in the same news article depended on the brand or outlet name (e.g., CNN or FOX) study participants saw.
  \\ \hline
\caption{Cognitive / Perception Bias} 
\label{table:categories4}
\end{longtable}

\newpage

\end{document}